\documentclass{article}

\usepackage{microtype}
\usepackage{graphicx}
\usepackage{subfigure}
\usepackage{booktabs} %

\usepackage{hyperref}

\usepackage[preprint]{icml2024}

\usepackage{amsmath}
\usepackage{amssymb}
\usepackage{mathtools}
\usepackage{amsthm}

\usepackage{xcolor}
\usepackage{tabularx}
\usepackage{subcaption}
\usepackage{multirow}
\usepackage{booktabs}
\usepackage{array}

\usepackage[capitalize,noabbrev]{cleveref}

\usepackage{subcaption}

\theoremstyle{plain}
\newtheorem{theorem}{Theorem}[section]

\theoremstyle{definition}
\newtheorem{definition}[theorem]{Definition}
\newtheorem{assumption}[theorem]{Assumption}
\theoremstyle{remark}

\usepackage[textsize=tiny]{todonotes}

\icmltitlerunning{\scheme{}: \capitalizedTitle{}}

\newcommand{\scheme}[0]{SEAL}
\newcommand{\capitalizedTitle}{Entangled White-box Watermarks on Low-Rank Adaptation}
\newcommand{\logp}[0]{ \(-\log(\text{p-value})\)}
\newcommand{\suppl}[0]{Appendix}

\newcommand{\prettyIndent}[0]{\\ \quad \quad \quad}

\newcommand{\smallstd}[1]{\(\scriptstyle{\pm{#1}}\)}

\newcommand{\secref}[1]{Sec.~\ref{#1}}      %
\newcommand{\figref}[1]{Fig.~\ref{#1}}      %
\newcommand{\tabref}[1]{Table~\ref{#1}}     %
\newcommand{\algref}[1]{Alg.~\ref{#1}}      %

\newcommand{\defref}[1]{Def.~\ref{#1}}      %

\begin{document}

\twocolumn[
\icmltitle{\scheme{}: \capitalizedTitle}

\icmlsetsymbol{equal}{*}

\begin{icmlauthorlist}
\icmlauthor{Giyeong Oh}{yyy}
\icmlauthor{Saejin Kim}{yyy}
\icmlauthor{Woohyun Cho}{yyy}
\icmlauthor{Sangkyu Lee}{yyy}
\icmlauthor{Jiwan Chung}{yyy}
\icmlauthor{Dokyung Song}{xxx}
\icmlauthor{Youngjae Yu}{yyy}
\end{icmlauthorlist}

\icmlaffiliation{yyy}{Department of Artificial Intelligence, Yonsei University, Seoul, Republic of Korea}
\icmlaffiliation{xxx}{Department of Computer Science and Engineering, Yonsei University, Seoul, Republic of Korea}

\icmlcorrespondingauthor{Dokyung Song}{dokyungs@yonsei.ac.kr}
\icmlcorrespondingauthor{Youngjae Yu}{youngjae4yu@gmail.com}

\icmlkeywords{Copyright Protection, Watermarking, DNN Watermarking, Machine Learning, Transfer Learning, ICML}

\vskip 0.3in
]

\printAffiliationsAndNotice{}  %

\begin{abstract}
Recently, LoRA and its variants have become the \textit{de facto} strategy for training and sharing task-specific versions of large pretrained models, thanks to their efficiency and simplicity. 
However, the issue of copyright protection for LoRA weights, especially through watermark-based techniques, remains underexplored.
To address this gap, we propose \textbf{\scheme} (SEcure wAtermarking on LoRA weights), the universal white-box watermarking for LoRA.
\scheme{} embeds a secret, non-trainable matrix between trainable LoRA weights, serving as a \emph{passport} to claim ownership.
\scheme{} then entangles the passport with the LoRA weights through training, without extra loss for entanglment, and distributes the finetuned weights after hiding the passport.
When applying \scheme{}, we observed no performance degradation across commonsense reasoning, textual/visual instruction tuning, and text-to-image synthesis tasks.
We demonstrate that \scheme{} is robust against a variety of known attacks: removal, obfuscation, and ambiguity attacks.
\end{abstract}

\section{Introduction}
Recent years have witnessed an increasing demand for protecting deep neural networks (DNNs) as intellectual properties (IPs), mainly due to the significant cost of collecting quality data and training DNNs on it.
In response, researchers have proposed various DNN watermarking methods for DNN copyright protection \cite{nagai2018digital,darvish2019deepsigns,zhang2018protecting,fan2019rethinking,zhang2020passport,xu2024hufu,lim2022protect}, which work by secretly embedding identity messages into the DNNs during training.
The IP holders can present the identity messages to a verifier in the event of a copyright dispute to claim ownership.

\begin{figure}[!t]
    \centering
    \includegraphics[width=\columnwidth]{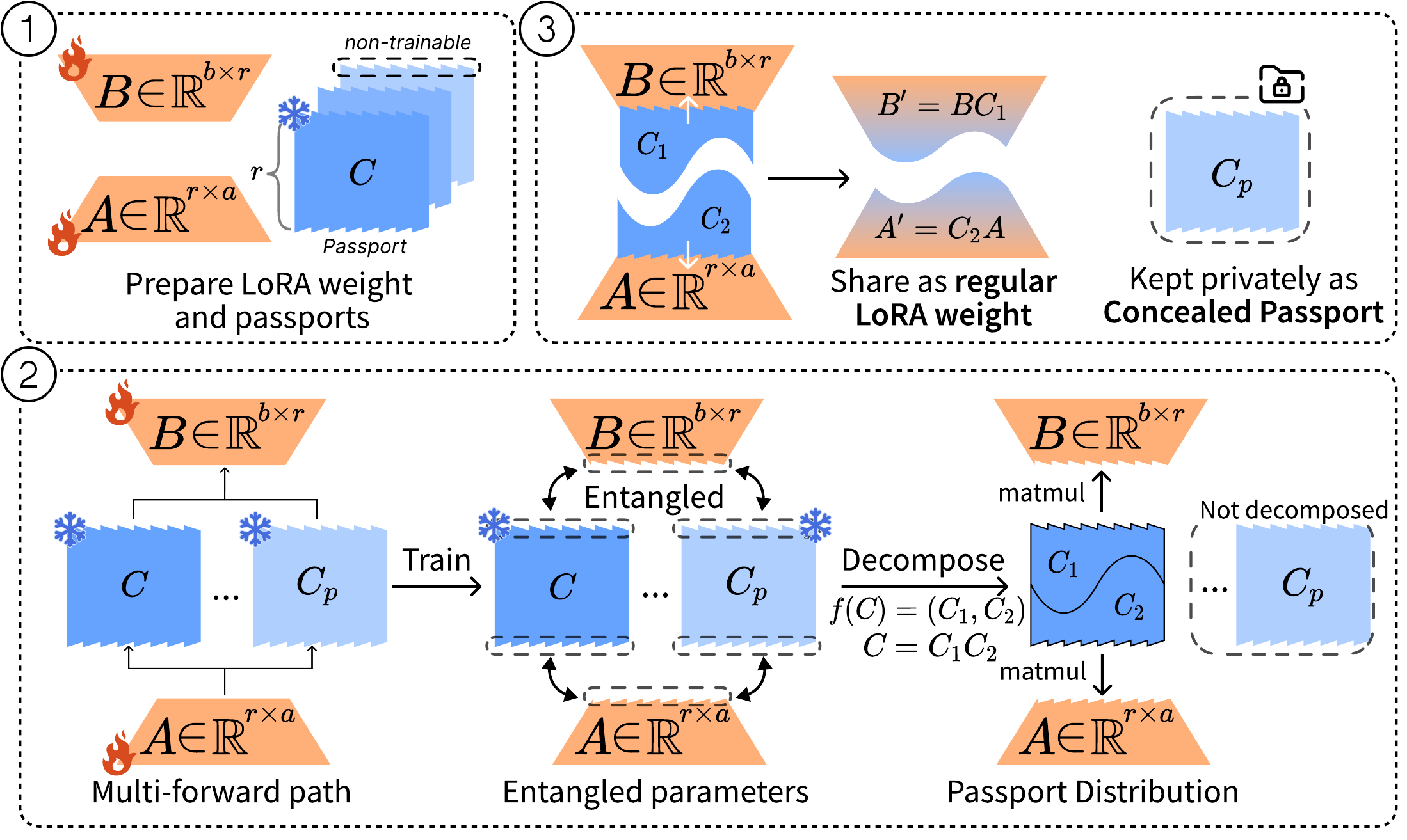}
    \caption{Overview of \scheme{}.
    (1) We begin with LoRA’s weights \({A}\) and \({B}\), plus non-trainable passports \({C}, {C}_p\).
    (2) During training, \({C}\) and \({C}_p\) are inserted between \({B}\) and \({A}\), forcing the model to rely on them and thus entangling the weights with the passports.
    (3) Afterward, \({C}\) is factorized via \(f({C})=({C}_1,{C}_2)\) and merged into \({B}\) and \({A}\), resulting in standard-looking LoRA weights \({B}'\) and \({A}'\). Meanwhile, \({C}_p\) remains private for ownership verification.}
    \label{fig:main_figure_1}
\end{figure}

Meanwhile, recent Parameter Efficient FineTuning (PEFT)~\cite{ding2023parameter} strategies, particularly Low-Rank Adaptation (LoRA)~\cite{hu2021lora}, 
have revolutionized how many domain-specific DNNs - especially Large Language Models (LLMs)~\cite{llama3modelcard,jiang2023mistral,team2024gemma,yang2024qwen2} and Diffusion Models (DMs)~\cite{Rombach_2022_CVPR} - are built and shared.
LoRA's efficacy stems from its lightweight adaptation layers, which introduce no additional inference overhead while preserving similar performance to fully fine-tuned models~\cite{zhao2024lora, jang2024lora, peft}. 
These qualities have led to a surge in open-source adaptation, as evidenced by more than 100k publicly shared LoRA weights on platforms such as \href{https://huggingface.co}{Hugging Face}, \href{https://civitai.com}{Civit AI}~\cite{luo2024stylus}.
In addition, several variants such as QLoRA~\cite{dettmers2024qlora}, LoRA+~\cite{hayoulora+}, and DoRA~\cite{liu2024dora} have emerged to further optimize resource usage and boost efficient domain adaptation. 
Due to these factors, LoRA's training framework has been established as the de facto approach in open-source communities for customizing large pretrained models to domain-specific tasks.

Although LoRA-based methods rely on pretrained foundation models, their uniquely trained adaptation weights themselves represent valuable IP that merits protection.
Unfortunately, existing white-box DNN watermarking schemes are not suitable for LoRA structure where weights are commonly released in open source,
as they only support embedding identity messages in specific architecture-bounded components, such as kernels in convolutional layers \cite{nagai2018digital,liu2021watermarking,zhang2020passport,lim2022protect}. %
By contrast, some approaches focus on utilizing LoRA to protect the original pretrained weights rather than safeguarding the LoRA itself~\cite{fengaqualora}, 
exposing these methods' failure to address LoRA's unique properties.

To address this gap, we propose \scheme{}, the universal watermarking scheme designed to protect the copyright of LoRA weights.
The key insight of \scheme{} is the integration of constant matrices, \emph{passports}, between the LoRA weight, acting as a hidden identity message that is difficult to extract, remove, modify, or even counterfeit, thus offering robust IP protection.
During fine-tuning, these passports naturally direct gradients through themselves, eliminating the need for additional constraint losses.
After fine-tuning, \scheme{} seamlessly decomposes the passport into two parts, each integrated into the up and down blocks, ensuring the final model is structurally indistinguishable from LoRA weights. %

We validate our \scheme{} against an array of attacks—removal \cite{han2015deepcompression}, obfuscation \cite{yan2023rethinking, pegoraro2024deepeclipse}, and ambiguity \cite{fan2019rethinking}—demonstrating that any attempt to remove or disrupt the passport severely degrades model performance. \scheme{} imposes no performance degradation on the host task; in many cases, it matches or even surpasses the fidelity of standard LoRA weights across various tasks.

In summary, our contributions are three-fold:
\begin{enumerate}
    \setlength{\itemsep}{3pt} %
    \setlength{\parskip}{0pt} %
    \setlength{\topsep}{0pt} %
    \item \textbf{Simple yet Strong Copyright Protection for LoRA.} We present \scheme{}, the universial watermarking scheme for protecting LoRA weights by embedding a hidden identity message using a constant matrix, \emph{passport}, eliminating the need for additional loss terms, offering a straightforward yet robust solution.
    \item \textbf{No Performance Degradation.}
    We demonstrate applying \scheme{} does not degrade the performance of the host task. 
    In practice, \scheme{} consistently achieves performance comparable to or even exceeding that of standard LoRA. 
    \item \textbf{Robustness Against Attacks.} We demonstrate \scheme{}'s resilience against various attacks, including removal, obfuscation, and ambiguity attacks, maintaining robust IP protection under severe adversarial conditions.
\end{enumerate}

\section{Preliminary}

\subsection{Low-Rank Adaptation}

LoRA \cite{hu2021lora} assumes that task-specific updates lie in a low-rank subspace of the model’s parameter space. 
It freezes the pretrained weights $W\in\mathbb{R}^{b \times a} $ and trains two low-rank matrices $ A\in\mathbb{R}^{r \times a} $ and $ B\in\mathbb{R}^{b \times r} $.  After training, the adapted weights are:
\begin{equation}
\label{eq_lora}
    W^{'} = W + \Delta W = W + BA
\end{equation}\
Because there are no activation functions between $A$ and $B$, one can simply add $BA$ to $W$ for efficient integration into the pretrained model.

\subsection{White-box DNN Watermarks}

White-box DNN watermarking techniques can be broadly categorized based on the \emph{location} of secret embedding or verification:
\begin{itemize}
    \setlength{\itemsep}{3pt} %
    \setlength{\parskip}{0pt} %
    \setlength{\topsep}{0pt} %
    
    \item \textbf{Weight-based.} These methods directly embed a secret bit sequence (e.g., \{\(+1\), \(-1\)\}) into the model \emph{parameters}. Verification often entails examining the trained weights to extract or validate the embedded bits~\cite{nagai2018digital, liu2021watermarking, fernandez2024functional}.

    \item \textbf{Activation-based.} Here, watermarks are embedded in the \emph{feature maps} of specific layers. By injecting specialized inputs, one can detect the hidden signature from the activations that uniquely respond to the watermark~\cite{darvish2019deepsigns, lim2022protect}.

    \item \textbf{Output-based.} These approaches ensure that the \emph{final output} from the model contains a watermark. Even in a white-box scenario, the verification is primarily conducted on the model’s output rather than its internal parameters~\cite{kirchenbauerreliability,fernandez2023stable,fengaqualora}.

    \item \textbf{Passport-based.} This line of work inserts an additional linear or normalization layer (\emph{passport layer}) into the model, so that using the correct passport yields normal performance, while invalid passports degrade the accuracy. During ownership verification, the legitimate passport is presented to confirm the model's fidelity, effectively distinguishing rightful owners from adversaries~\cite{fan2019rethinking, zhang2020passport}.
\end{itemize}
Unlike weight-, activation-, or output-based methods, passport-based watermarking ties model performance to hidden parameters (\emph{passports}). It does not require special triggers or depend solely on model outputs. Instead, ownership is verified by a passport that restores high accuracy, securely linking model weights and the embedded secret.

\subsection{Threat Model and Evaluation Criteria}
\label{sec:threat_model_preliminary}

\paragraph{Attack Types.}
Under a white-box setting, adversaries are assumed to have full access to the model weights. We idntify three primary attack categories: 

\textbf{(1) Removal}, where attackers prune unimportant parameters \cite{lecun1989optimal,han2015deepcompression,nagai2018digital,darvish2019deepsigns}, finetune the model without watermark constraints \cite{chen2021refit,guo2021finetuning}.

\textbf{(2) Obfuscation}, where attackers restructure the architecture to disrupt watermark extraction \cite{yan2023rethinking,pegoraro2024deepeclipse,li2023linear}, all while preserving model functionality equivalently.

\textbf{(3) Ambiguity}, where counterfeit keys or watermarks are forged to deceive verifiers into believing the adversary is the rightful owner \cite{fan2019rethinking,zhang2020passport,chen2023effective}.

\textbf{Adversary Assumptions.}
We assume the adversary obtains the open-sourced weights but \textbf{lacks} have access to the original finetuning data and \textbf{is unable} to retrain from scratch. Consequently, they aim to preserve the model's performance, as excessive degradation would nullify its value. While they know our watermarking \emph{scheme} (Kerckhoff's principle), the secret key itself remains undisclosed.

\textbf{Evaluation Criteria.}
A robust DNN watermark should satisfy two requirements \cite{nagai2018digital}:
\textbf{Fidelity}, meaning the watermark does not degrade the model's original performance;
and \textbf{Robustness}, ensuring the watermark resists removal, obfuscation, and ambiguity attacks.

\begin{algorithm}[!tb]
\caption{\scheme{} Training Procedure}
\begin{algorithmic}
\STATE {\bfseries Input:} Pretrained weights $W$, LoRA rank $r$ \prettyIndent Passports $C, C_p$ \prettyIndent Training dataset $\mathcal{D}$, Epochs $E$
\STATE {\bfseries Output:} Public LoRA weights $B', A'$ \prettyIndent \; \, Private parameters $B, A, C, C_p$

\STATE Initialize $A \in \mathbb{R}^{r \times a}$, $B \in \mathbb{R}^{b \times r}$ as trainable parameters.
\STATE Set $C, C_p \in \mathbb{R}^{r \times r}$ as non-trainable passports.
\FOR{$e = 1$ {\bfseries to} $E$}
   \FOR{each batch $(x,y)$ in $\mathcal{D}$}
       \STATE Randomly pick $C$ or $C_p$
       \STATE Compute $W' = W + BCA$ or $W' = W + B C_p A$
       \STATE Compute $\mathcal{L}(W', x, y)$
       \STATE Backpropagate $\nabla \mathcal{L}$
   \ENDFOR
\ENDFOR

\STATE Decompose $C$ into $C_1, C_2$ where $C = C_1 C_2$
\STATE Set $B' = B C_1$ and $A' = C_2 A$.

\STATE \textbf{return} $B', A'$ (Public), and keep $B, A, C, C_p$ (Private)
\end{algorithmic}
\label{alg:training}
\end{algorithm}

\section{\scheme{}: The Watermarking Scheme}

\begin{figure}[t]
    \centering
    \includegraphics[width=0.95\linewidth]{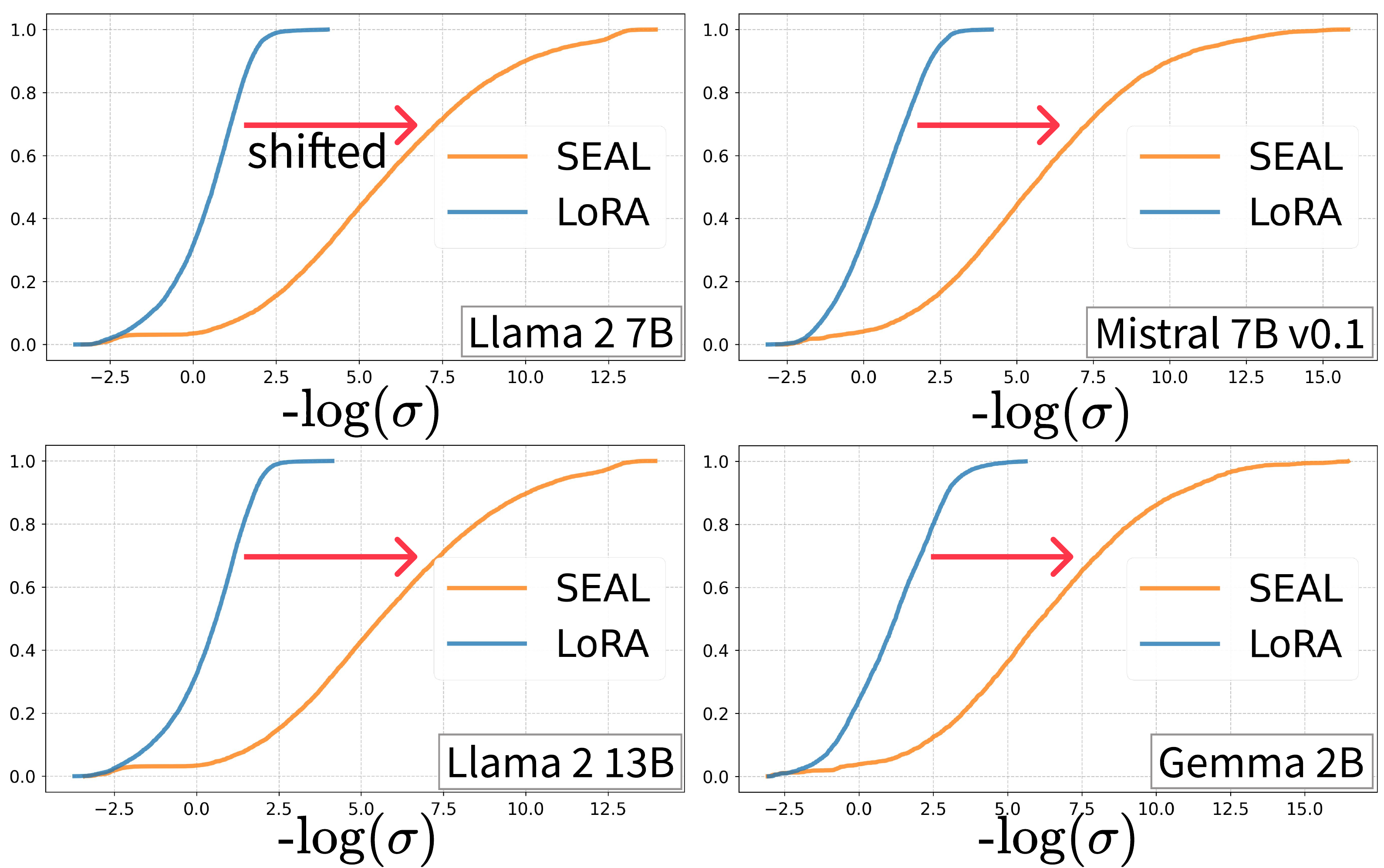}
    \caption{
    Negative log singular value (CDF), collection of top-32 singular values. LoRA (blue) vs.\ \scheme{} (orange) 
    across Llama-2, Mistral, and Gemma models.
    }
    \label{fig:svd_logvalues}
\end{figure}

For clarity, the symbols used throughout this section are listed in~\tabref{tab:notation}.

\subsection{Impact of the Constant Matrix between LoRA}

In \figref{fig:svd_logvalues}, we compare the distributions of 
negative log singular values, \(-\log(\sigma)\),
from a standard LoRA model 
, \(\mathbb{N}(B,A)\), against our proposed \scheme{}, \(\mathbb{N}(B,A,C)\), approach on multiple models: Llama-2-7B/13B~\cite{touvron2023llama}, 
Mistral-7B-v0.1~\cite{jiang2023mistral},
and Gemma-2B~\cite{team2024gemma}. 
For each trained model, 
we reconstruct the learned weight \(\Delta W\), 
collect the top-32 singular values \(\sigma\) from each module, 
and plot \(-\log(\sigma)\) in a cumulative distribution function (CDF).

We observe that the \scheme{} curves systematically shift to the right compared to LoRA. 
This shift implies that the learned subspace under \scheme{} is more evenly spread across multiple singular directions, rather than being dominated by just a few large singular values. 
Such broad coverage in the singular spectrum can bolster robustness: 
altering or removing the watermark in one direction has a limited effect, 
as the watermark is “spread out” in multiple directions.
Further gradient-based analyses are provided 
in \suppl{}~\ref{appendix:training_procees_seal}.

\subsection{Comparison with Existing Passport Methods}
Unlike prior passport-based methods~\cite{fan2019rethinking,zhang2020passport} 
that typically introduce an additional loss term (a regularization or constraint to embed the passport) \emph{and} keep the passport layer trainable,
\scheme{} employs a \emph{non-trainable} matrix $C$ inserted directly into LoRA’s block, eliminating the need for auxiliary loss terms. 
Consequently, our approach differs from existing methods 
on \emph{two} key fronts—no extra loss and a non-trainable passport—%
making a one-to-one comparison problematic.

\subsection{Entangling Passports during Training}
\label{sec:entangle_passport}

\scheme{} embeds the watermark during training by inserting the non-trainable, constant matrix $C$ between the trainable parameters $B$ and $A$. Doing so effectively \emph{entangles} the given passport with $B$ and $A$.
The concept of entanglement is superficially similar to the entanglement proposed by \citeauthor{jia2021entangled}
It involves indistinguishable distributions between host and watermarked tasks. In our context, we define entanglement as follows.

\begin{definition}[Entanglement]
\label{def_self_entanglement}
Given trainable parameters $A$ and $B$, and a non-trainable parameter $C$, $A$ and $B$ are in \emph{entanglement} via $C$ \textit{if and only if} they produce the correct output for the host task when $C$ is present between them.
\end{definition}

As despicted~\algref{alg:training}, $C$ directly influences the computations of $B$ and $A$ during the forward pass, and modifies the gradient flow in the backward pass, thereby embedding itself through a normal training process.
The IP holder incorporates both \(C\) and \(C_p\) during training, alternating them according to the batch size.

\subsection{Hiding Passport for Distribution} 
\label{sec:distribution}

After successfully establishing the entanglement between the passport and other trainable parameters, the passport must be hidden before distribution.
Therefore, we decompose the passport, \( C \), of the IP holder into two matrices such that their product reconstructs \( C \), as shown in \figref{fig:main_figure_1}.
\begin{definition}[Decomposition Function]
\label{def:decomposition_function}
    For a given constant \( C \), a function \( f \) is a decomposition function of \( C \) where $f: C \mapsto (C_1, C_2)$ such that \( C_{1}C_{2} = C \).
\end{definition}
The decomposition function ensures that models trained with \scheme, which contain three matrices per layer, $ \mathbb{N}(B, A, C)$, 
can be distributed in a form that resembles standard LoRA implementations with only two matrices, $ \mathbb{N}(B', A')$. 
In the decomposition process, the IP holder can camouflage the passport $C$ within the open-sourced weight by distributing its decomposed components into $B$ and $A$.

An example of decomposition using SVD is
\[ 
    f_{svd}(C) = (U_{C} \sqrt{\Sigma_{C}}, \sqrt{\Sigma_{C}} V_{C}^T),
\]
where \( C = U_{C} \Sigma_{C} V_{C}^T \).
Using SVD decomposition function, $f_{svd}$, the resulting component of $\mathbb{N}(B', A')$ is 

\[
    B' 
    =
    B \, (U_{C} \sqrt{\Sigma_{C}}) \; \text{and} \; A' = (\sqrt{\Sigma_{C}} V_{C}^T) \, A.
\]
We will use $f_{svd}$ as the default decomposition function unless otherwise specified.
Notably, $C_p$ is not distributed into either $B$ or $A$.

\subsection{Extraction on Embedded Passport}

To extract the embedded passport from LoRA converted SEAL weight, $\mathbb{N}(B', A')$, we have to assume that $A$ and $B$, which are trained SEAL weights, are full rank matrices. 

\begin{assumption}[Rank of trained \scheme{} weights]
\label{assumption_pseudo}
Trained \scheme{} weights $B$ and $A$ are full rank matrices with $r$.
\end{assumption}
By Assumption \ref{assumption_pseudo}, 
$A$ and $B$ have the pseudo-inverse $A^{\dagger}$, $B^{\dagger}$ such that $ A A^{\dagger} = I_{r}$, $B^{\dagger} B = I_r$ where $I_r \in \mathbb{R}^{r \times r}$ is the identity matrix.
As shown in~\algref{alg:extraction}, the method for extracting the passport from $B'A'$ is multiplying $A^{\dagger}$, $B^{\dagger}$ in the right/left side of $B'A'$, respectively. Thus, only the legitimate owner, who has original \scheme{} weights $B$ and $A$, can extract the concealed passport, $C$, from $\mathbb{N}(B', A')$. 

\subsection{Passport-based Ownership Verification}

\subsubsection{Extraction}

\begin{algorithm}[tb]
\caption{\scheme{} Verification by Extraction}
\label{alg:extraction}
\begin{algorithmic}
\STATE {\bfseries Input:} Public weights $(A', B')$, \\ \quad \quad \quad Claimant submits $(A, B, C)$
\STATE {\bfseries Output:} True or False
\STATE Compute $C_{ext}=B^\dagger B' A' A^\dagger$.
\IF{$C_{ext}$ $\approx$ $C$ (statistically) } 
    \STATE \textbf{return} True \hfill // Claimant passes

\ELSE
    \STATE \textbf{return} False \hfill // Claimant fails extraction
\ENDIF
\end{algorithmic}
\end{algorithm}

\citeauthor{yan2023rethinking} demonstrate that it is possible to neutralize the extraction process of watermarking schemes by altering the distribution of parameters while maintaining functional invariance. Given that the adversary is aware of \scheme{} and we assume a white-box scenario, the adversary could generate the triplet $\widetilde{A}, \widetilde{B}, \widetilde{C}$ for the verifier during the extraction process such that
\[
    \text{rank}(\widetilde{A}) = r \, \text{ and } ,\ \widetilde{B} = B'A'\widetilde{A}^{\dagger}\widetilde{C}^{\dagger}
\]
In this process, even if $\widetilde{C} \neq C$, which is the truly distributed passport, the verifier could be confused about who the legitimate owner is.
For this reason, extraction should only be used when the legitimate owner is attempting to verify whether their passport is embedded in a suspected model.
It should not be relied upon in scenarios where a third-party verifier is required for a contested model, as it is vulnerable in such cases. 

Therefore, it is crucial to leverage the inherent characteristic of passport-based schemes—where performance degradation occurs if the correct passport is not presented—allowing a third-party verifier to determine the legitimate owner accurately~\cite{fan2019rethinking}.

\subsubsection{Measuring Fidelity}
\label{subsec:fidelity_score}

Recall that \scheme{} entails two passports, $(C, C_p)$, both entangled with the LoRA weights $(B,A)$. 
To gauge how similarly these two passports preserve the model’s performance, we define a \emph{fidelity gap}:
\[
    \epsilon_{T}
    =
    \Bigl|
        M_{T}\bigl(\mathbb{N}(B,A,C)\bigr)
        -
        M_{T}\bigl(\mathbb{N}(B,A,C_p)\bigr)
    \Bigr|,
\]
where $M_{T}$ is the task-specific metric for the adaptation layer $\mathbb{N}(B,A,\cdot)$ on task $T$. 
A small $\epsilon_{T}$ indicates that $C$ and $C_p$ yield near-identical performance, 
implying they were jointly entangled during training.
By contrast, if two passports are \emph{not} entangled with $(B,A)$, switching between them would degrade the model’s accuracy, producing a large $\epsilon_{T}$.

In a legitimate setting, the owner’s $(C, C_p)$ should incur almost no performance difference ($\epsilon_{T}$ close to zero). 
An attacker forging a second passport, however, cannot maintain the same fidelity gap without retraining the entire LoRA model. 
Hence, $\epsilon_{T}$ naturally serves as a verification criterion for rightful ownership. 
Detailed formulation is in \secref{sec:verification}

\newcommand{\seald}{SEAL$^{\dagger}$ (Ours)}
\newcommand{\sealdd}{SEAL$^{\dagger}$}
\newcommand{\seal}{SEAL (Ours)}

\begingroup
\setlength{\tabcolsep}{1.8pt} %
\renewcommand{\arraystretch}{0.8} %

\begin{table*}[!h]
\centering
\caption{Commonsense Reasoning Performance (3-run Avg.). Scores are averaged over three random seeds, with standard deviation shown in a smaller font for the last column. \sealdd{} denotes using a constant matrix $C$ from normal distribution.}

\begin{tabular}{llccccccccc}
\toprule
                   & \textbf{Method}   & \textbf{BoolQ} & \textbf{PIQA}  & \textbf{SIQA}  & \textbf{HellaSwag} & \textbf{Wino.}  & \textbf{ARC-e} & \textbf{ARC-c} & \textbf{OBQA}  & \textbf{Avg.} ↑ \\
\midrule
\multirow{3}{*}
{LLaMA-2-7B}       & LoRA      & 73.75 & 82.99 & 79.85 & 86.14     & 85.06 & 86.15 & 73.63 & 85.80 & 81.67 \smallstd{1.03} \\
                   & \seal     & 72.70 & 85.27 & 81.27 & 90.15     & 85.79 & 87.07 & 74.60 & 85.00 & 82.73 \smallstd{0.14} \\
                   & \seald    & 73.19 & 86.31 & 81.95 & 91.21     & 86.69 & 88.55 & 75.51 & 86.80 & \textbf{83.78} \smallstd{0.27} \\
\midrule
\multirow{3}{*}
{LLaMA-2-13B}      & LoRA      & 75.57 & 86.98 & 81.39 & 91.82     & 88.53 & 90.08 & 78.78 & 86.67 & 84.98 \smallstd{0.17} \\
                   & \seal     & 75.34 & 87.41 & 83.28 & 93.33     & 88.42 & 90.68 & 79.61 & 86.73 & 85.60 \smallstd{0.34} \\
                   & \seald    & 75.67 & 88.63 & 83.21 & 93.95     & 89.29 & 91.72 & 81.46 & 88.53 & \textbf{86.56} \smallstd{0.10} \\
\midrule
\multirow{3}{*}
{LLaMA-3-8B}       & LoRA      & 74.76 & 88.22 & 80.96 & 92.00     & 86.08 & 90.09 & 82.41 & 86.30 & 85.10 \smallstd{1.39} \\
                   & \seal     & 73.88 & 88.23 & 82.29 & 94.84     & 88.35 & 91.67 & 82.00 & 86.27 & 85.94 \smallstd{0.29} \\
                   & \seald    & 75.78 & 90.37 & 83.25 & 96.05     & 89.92 & 93.49 & 84.73 & 90.60 & \textbf{88.02} \smallstd{0.11} \\
\midrule
\multirow{3}{*}
{Gemma-2B}         & LoRA      & 67.05 & 83.19 & 77.26 & 87.07     & 79.74 & 83.91 & 69.34 & 79.87 & 78.43 \smallstd{0.32} \\
                   & \seal     & 66.56 & 81.79 & 77.65 & 84.82     & 79.16 & 82.79 & 68.40 & 79.20 & 77.55 \smallstd{0.04} \\
                   & \seald    & 66.70 & 82.50 & 78.88 & 87.57     & 80.19 & 83.81 & 69.97 & 79.87 & \textbf{78.68} \smallstd{0.11} \\
\midrule
\multirow{3}{*}
{Mistral-7B-v0.1}  & LoRA    & 75.92 & 90.72 & 81.78 & 94.68     & 88.69 & 93.10 & 83.36 & 88.30 & 87.07 \smallstd{0.27} \\
                   & \seal   & 73.08 & 87.52 & 81.92 & 91.23     & 87.97 & 90.19 & 78.70 & 88.13 & 84.84 \smallstd{0.44} \\
                   & \seald  & 76.92 & 90.42 & 82.51 & 94.57     & 90.08 & 93.31 & 83.25 & 91.73 & \textbf{87.85} \smallstd{0.02} \\
\bottomrule
\end{tabular}
\label{table:commonsense}
\end{table*}

\endgroup

\subsubsection{Verification}
\label{sec:verification}

\begin{algorithm}[tb]
   \caption{\scheme{} Verification by Fidelity}
   \label{alg:fidelity}
\begin{algorithmic}
\STATE \textbf{Input:} Suspected $(B',A')$; \prettyIndent{} Claimant submits $(B,A,C_a,C_b)$; \prettyIndent{} Threshold $\epsilon_T$; Task $T$; Measurement $M_T$
\STATE \textbf{Output:} True or False

\STATE // 1) Check if claimant’s parameters reconstruct $(B',A')$
\IF{$B\,C_{a}\,A \;=\; B'A'$}
   \STATE // 2) Evaluate fidelity gap
   \STATE $\Delta \leftarrow \bigl|\,M_T(\mathbb{N}(B,A,C_a)) - M_T(\mathbb{N}(B,A,C_b))\bigr|$
   \IF{$\Delta \le \epsilon_T$}
        \STATE \textbf{return} True \hfill // Ownership verified
   \ELSE
        \STATE \textbf{return} False \hfill // Passport validation failed
   \ENDIF
\ELSE
    \STATE \textbf{return} False \hfill // Parameters mismatch
\ENDIF
\end{algorithmic}
\end{algorithm}

The fundamental idea behind passport-based watermarking is that any \emph{forged} passport significantly degrades the model’s performance \cite{fan2019rethinking}, resulting in a fidelity gap $\Delta > \epsilon_T$. 
As shown in~\algref{alg:fidelity}, the suspected model $(B',A')$ is first checked against the claimant’s $(B,A,C_a)$ to ensure they reconstruct the same adaptation weights. 
If so, we measure $\Delta$ between $C_a$ and $C_b$ (the two passports) via the task metric $M_T$. \defref{def:verification} then concludes that ownership is verified if and only if $\Delta \le \epsilon_T$:

\begin{definition}[Verification Process]
\label{def:verification}
Assume $\mathbb{N}(B',A')=\mathbb{N}(B,A,C_a)\neq\mathbb{N}(B,A,C_b)$. Define
\[
\Delta := \bigl|\,M_T(\mathbb{N}(B,A,C_a)) - M_T(\mathbb{N}(B,A,C_b))\bigr|,
\]
\[
V\bigl(M_T, \epsilon_T, \mathbb{N}(B,A,\cdot)\bigr) \;=\;
\begin{cases}
\text{True}, & \text{if } \Delta \le\epsilon_T,\\
\text{False}, & \text{otherwise}.
\end{cases}
\]
\end{definition}

In practice, the legitimate owner can submit $(B,A,C,C_p)$, achieving $\Delta \le \epsilon_T$. 
Any adversary forging $(\widetilde{C}, \widetilde{C}_{p\text{-adv}})$ lacks the entanglement from training, 
and fails to keep $\Delta$ within the threshold. 
See \suppl{}~\ref{appendix:forge_multi_passports} for an extended discussion.

\section{Experiments}
\label{sec:experiments}
\subsection{Experimental Setup}

\subsubsection{Fidelity}
To demonstrate that the performance of models after embedding \scheme{} passports does not degrade, we conducted experiments across both language and image modalities. Initially, we evaluate our model by comparing it with various open-source Large Language Models (LLMs) such as LLaMA-2-7B/13B \cite{touvron2023llama}, LLaMA-3-8B \cite{llama3modelcard}, Gemma-2B \cite{team2024gemma}, and Mistral-7B-v0.1 \cite{jiang2023mistral} on commonsense reasoning tasks. Next, we verify the model's effectiveness on instruction tuning tasks. Following this, we extend our approach to the Vision Language Model (VLM) \cite{liu2024llava} by evaluating the model's performance on visual instruction tuning. Finally, we assess \scheme{}'s capabilities on image-generative tasks \cite{Rombach_2022_CVPR}.

\subsubsection{Robustness}
We evaluated the robustness of \scheme{} against removal, obfuscation and ambiguity attacks by evaluating fidelity scores in commonsense reasoning tasks. For removal and obfuscation attacks, the presence of the extracted watermark was confirmed through hypothesis testing. For ambiguity attacks, fidelity scores were used to verify genuine versus counterfeit passports, as defined in \defref{def:verification}.

\label{exp:fidelity}

\subsection{Commonsense Reasoning Task}
\label{exp:commonsense_reasoning}
\tabref{table:commonsense} presents the performance comparison across commonsense reasoning tasks: \textbf{BoolQ}~\cite{clark2019boolq}, \textbf{PIQA}~\cite{bisk2020piqa}, \textbf{SIQA}~\cite{sap2019socialiqa}, \textbf{HellaSwag}~\cite{zellers2019hellaswag}, \textbf{Wino.}~\cite{sakaguchi2021winogrande}, \textbf{ARC-e}, \textbf{ARC-c}~\cite{clark2018think}, and \textbf{OBQA}~\cite{OpenBookQA2018}. The dataset combines multiple sources, as detailed in \cite{hu-etal-2023-llm}. We train LLMs on 3-epochs on the combined dataset.
The experimental results emphasize that \scheme{} can be seamlessly integrated into existing LoRA architectures, 
without affecting performance degradation.

\begin{table}[!t]
    \setlength{\tabcolsep}{1.8pt} %
    \renewcommand{\arraystretch}{1.0} %
    \centering
    \caption{ Fidelity across various tasks involves \textbf{Inst. Tune} (instruction tuning), \textbf{MT-B} (MT-Bench) and t2i task. Visual \textbf{Inst. Tune} score averages over seven vision-language tasks (see \suppl{}). CLIP-I and DINO demonstrate subject fidelity scores, while CLIP-T shows prompt fidelity scores. 
    }

    \begin{tabular}{l|cc|ccc}
    \toprule
\multirow{2}{*}{Task} 
    & \multicolumn{2}{c|}{\textbf{Inst. Tune}} & \multicolumn{3}{c}{\multirow{2}{*}{Text-to-Image}} \\
    & \multicolumn{1}{c}{Textual} & \multicolumn{1}{c|}{Visual}  & && \\
    \midrule
      Metric $\uparrow$ & MT-B & Acc. & CLIP-T & CLIP-I & DINO. \\ 
        \midrule
        LoRA & \textbf{5.83}  & \textbf{66.9}  & \textbf{0.20} & \textbf{0.80} & \textbf{0.68}  \\
        SEAL & 5.81  & 63.1  & \textbf{0.20} & \textbf{0.80} & 0.67  \\
    \bottomrule
    \end{tabular}
    \label{tab:it_vit_t2i_all_in_one}
\end{table}

\subsection{Textual Instruction Tuning}
\label{exp:instruction_tuning}

\tabref{tab:it_vit_t2i_all_in_one} shows the scores for LLaMA-2-7B, instruction tuned with both LoRA and \scheme{}, using Alpaca dataset \cite{alpaca} with 3-epochs. The scores are averaged ratings given by \texttt{gpt-4-0613} on a scale of 1 to 10 for the models' responses to questions from MT-Bench \cite{zheng2023judging}.
Since the Alpaca dataset is optimized for single-turn interactions, the average score for single-turn performance from MT-Bench is used. 
The results indicate that \scheme{} achieves performance comparable to LoRA, thereby confirming its fidelity.

\subsection{Visual Instruction Tuning}
\label{exp:visual_instruction_tuning}

\tabref{tab:it_vit_t2i_all_in_one} shows the average performance across seven visual instruction tuning benchmarks \cite{goyal2017vqav2,hudson2019gqa,gurari2018vizwiz,lu2022sqa,singh2019vqat,li2023pope,liu2023mmbench} for LoRA and \scheme{} on LLaVA-1.5~\cite{liu2024llava} with detailed elaboration in \suppl{}~\ref{appendix:visit_results}.
As shown in~\tabref{tab:it_vit_t2i_all_in_one}, the performance of \scheme{} is comparable to that of LoRA.

\subsection{Text-to-Image Synthesis}
\label{exp:t2i_synthesis}

The experimentation with the Stable Diffusion model~\cite{Rombach_2022_CVPR} in conjunction with the dataset of DreamBooth~\cite{ruiz2023dreambooth} trained with LoRA
elucidates the versatility 
SEAL when integrated into diverse architectures. 
\tabref{tab:it_vit_t2i_all_in_one} provides a detailed comparison of subject fidelity, CLIP-I~\cite{pmlr-v139-radford21a}, DINO. \cite{caron2021emerging}, and prompt fidelity, CLIP-T, using the methods employed in \cite{nam2024dreammatcher}.
Our results confirm that \scheme{} maintains high fidelity and prompt accuracy without any degradation in model performance. 

\begin{figure*}[t]
    \centering\
    \includegraphics[width=\textwidth]{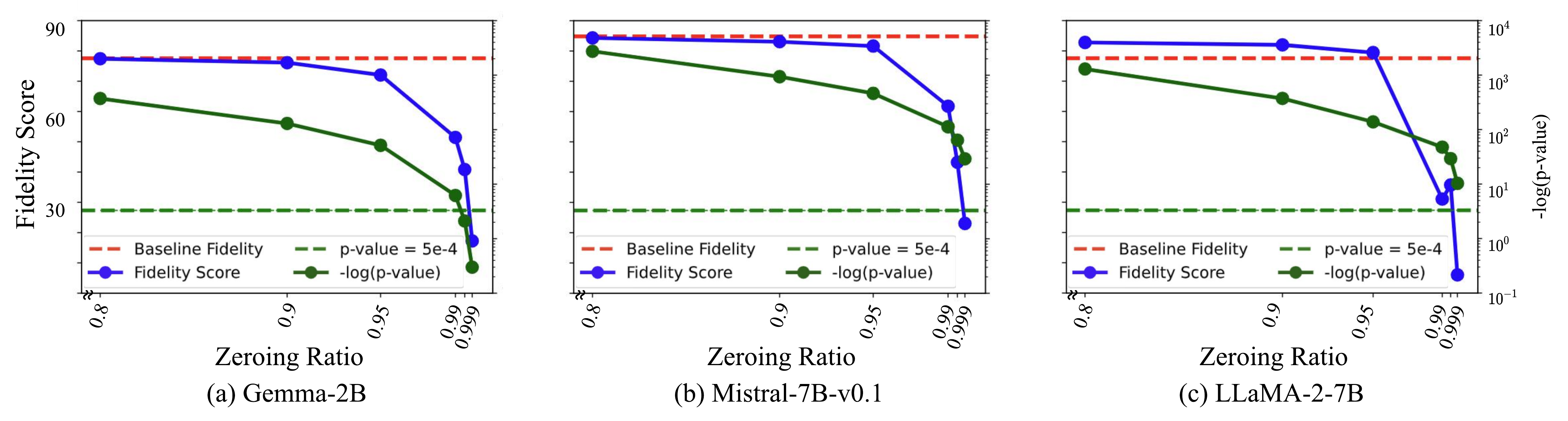}
    \caption{Pruning Attack. The x-axis represents the zeroing ratio of the smallest parameters of $\mathbb{N}(B',A')$ based on their L1 norms, the left y-axis shows the fidelity score on commonsense reasoning tasks, and the right y-axis displays the \logp{} on a log scale. If \logp{} is \emph{above} 3.3 (i.e., p-value $<5\times 10^{-4}$), detecting the watermark succeeds. The graphs show that as the zeroing ratio increases, the fidelity score decreases.
    This indicates the watermark remains detectable until \textbf{99.9\%} of the weights are zeroed, which significantly degrades the host task's performance.
    }
    \label{fig:pruning_attack}
\end{figure*}

\subsection{Integrating with LoRA Variants}
\begin{table}[!ht]
    \centering
    \caption{Average Commonsense Reasoning Performance on Llama-2-7B for LoRA, DoRA, and \scheme{}. 
    The notation \scheme{}+DoRA signifies that the \scheme{} approach has been applied in conjunction with the DoRA variant.
    Hyperparameter settings are in \suppl{}~\ref{appendix:training_details}.}
    \label{tab:dora+seal}
    \begin{tabular*}{0.85\linewidth}{l|c|c}
        \toprule
        \textbf{Method} & \textbf{Wall Time (h)} & \textbf{Avg.} \\
        \midrule
        LoRA &  12.0 & 81.67 \smallstd{1.03}  \\
        DoRA &  18.5 & 81.98 \smallstd{0.26} \\
        \scheme{} &  19.6 & \textbf{83.78} \smallstd{0.27}\\
        \scheme{} + DoRA &  27.8 & 81.88 \smallstd{1.08}\\
        \bottomrule
    \end{tabular*}
\end{table}

Thanks to its flexible framework, \scheme{} can easily be applied to a wide variety of LoRA variants.
In \tabref{tab:dora+seal}, we use DoRA~\cite{liu2024dora} as a case study to demonstrate that \scheme{} can seamlessly integrate with diverse LoRA-based methods, as exemplified by \scheme{}+DoRA. 

Even without any hyperparameter optimization, \scheme{} +DoRA matches accuracy of DoRA, highlighting that these variants can coexist with \scheme{} in a single pipeline without interference.
Further details on how \scheme{} applies to other LoRA variants, including \texttt{matmul}-based and other multiplicative approaches~\cite{edalati2022krona,hyeon2021fedpara}, can be found in \suppl{}~\ref{appendix:matmul_variants} and \ref{appendix:generalized_operators}.

\begin{table}[!ht]
    \centering
    \caption{Finetuning Attack. The detectability of passport on \scheme{} across either the same or different datasets. 
    }
    \setlength{\tabcolsep}{4pt} %
    \begin{tabular*}{0.76\linewidth}{l|cc|c}
        \toprule
        \textbf{Tasks} & \textbf{Acc.} & \textbf{MT-B} & \textbf{p-value} \\
        \midrule
        $\text{C}_{3e}$ & 83.1 & - & - \\
        $\text{I}_{3e}$ & - & 5.81 & - \\
        \midrule
        $\text{I}_{3e} \rightarrow \text{C}_{1e}$ & 60.2 & 4.94 & $1.71 \times 10^{-1171}$ \\
        $\text{C}_{3e} \rightarrow \text{I}_{1e}$ & 0.24 & 3.56 & $2.81 \times 10^{-178}$ \\
        $\text{C}_{3e} \rightarrow \text{C}_{1e}$ & 82.9 & - & $3.86 \times 10^{-3111}$ \\
        $\text{I}_{3e} \rightarrow \text{I}_{1e}$ & - & 3.78 & $9.08 \times 10^{-6}$ \\
        \bottomrule
    \end{tabular*}
 
    \label{tab:finetuning_another_datset}
\end{table}

\subsection{Pruning Attack}

\label{exp:pruning_attack}
Pruning attacks were performed on trained \scheme{} weights by zeroing out $\mathbb{N}(B', A')$ based on their L1 norms. And we extract passport, $C$, on pruned weight.
We used statistical testing instead of Bit Error Rate (BER) because, unlike prior work \cite{nagai2018digital,fernandez2024functional,zhang2020passport,fengaqualora} that used a small number of bits, $N{\sim}10^2$, the amount of our passport bits is approximately $N{\sim}10^5$, necessitating a different approach.
In hypothesis testing, if the p-value is smaller than our significance level ($\alpha$ = 0.0005), we reject the null hypothesis, \textit{the extracted watermark is an irrelevant matrix with $C$.} Rejecting the hypothesis implies that the extracted watermark is not random noise but exists within the model.

\figref{fig:pruning_attack} illustrates the fidelity score and \logp{} obtained by zeroing the 
smallest
parameters of $\mathbb{N}(B',A')$,
based on L1 norms.
The results demonstrate that removing the watermark necessitates zeroing \textbf{99.9\%} of the weights, which severely impacts the host task's performance, thereby confirming \scheme{}'s robustness against pruning attacks.

\begin{figure*}[!t]
    \centering
    \includegraphics[width=0.9\textwidth]{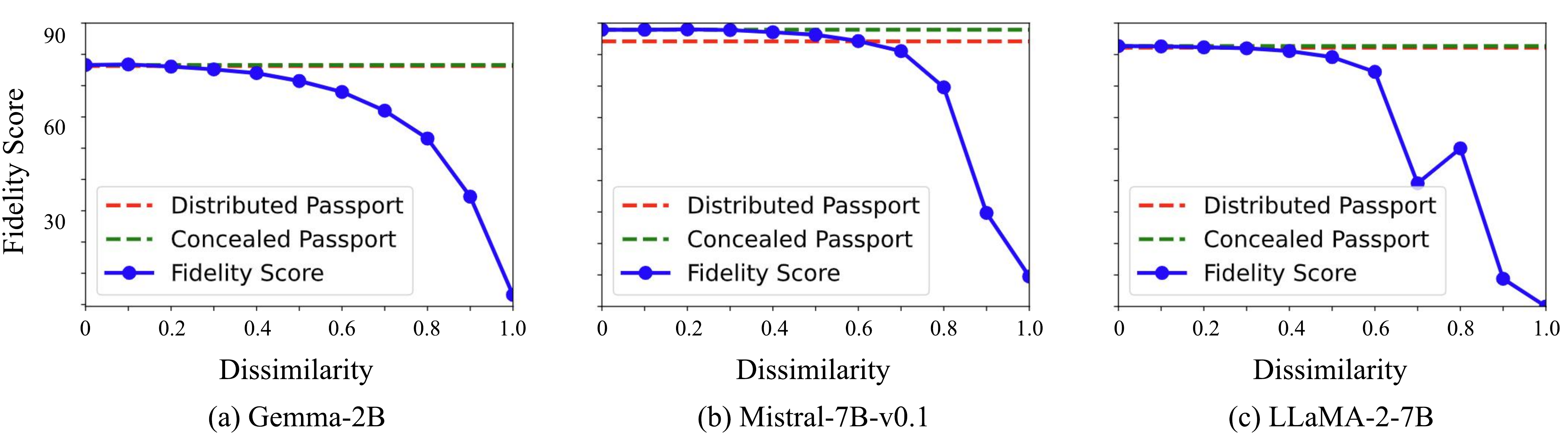}

    \caption{Ambiguity Attacks. Fidelity score, $M_{T}(\mathbb{N}(A, B, C_{t})$, as average accuracy on Commonsense Reasoning tasks, $T$, with the passport $C_{t}$, which is the inference time passport. The x-axis represents the dissimilarity, \( \gamma \), where \( C_{t} = (1 - \gamma)C_{p} + \gamma \widetilde{C}_{p\text{-adv}} \). \( {C}_{p} \) is the concealed passport, and \( \widetilde{C}_{p\text{-adv}} \) is the adversary' matrix. When \( \gamma > 0.6 \), the difference between fidelity scores significantly drops below the threshold of the verification process, \( \epsilon_{T} \), as shown in Table \ref{tab:passport_perfomance}.}
    \label{fig:ambiguity_attack}
\end{figure*}

\subsection{Finetuning Attack}
\label{exp:finetuning_attack}

In this experiment, we aimed to assess the robustness of \scheme{}'s watermark under finetuning attacks. The notation \( T_{ne} \) represents a task \( T \) fine-tuned for \( n \) epochs. Specifically, we resumed training on \scheme{} weights, $\mathbb{N}(B',A')$, that had been trained for 3 epochs on two tasks: commonsense reasoning ($\text{C}_{3e}$) and instruction tuning with Alpaca dataset \cite{alpaca} ($\text{I}_{3e}$). The notation $\text{T}_{3e} \rightarrow \text{T}'_{1e}$ represents post-finetuning with the respective dataset for 1 additional epoch using standard LoRA training, where $A'$ and $B'$ are the trainable parameters.

These finetuning scenarios were designed to simulate an adversarial attack, where the model is fine-tuned either on the original or a different dataset, such as finetuning on Alpaca for one epoch ($\rightarrow \text{I}_{1e}$) or on commonsense reasoning for one epoch ($\rightarrow \text{C}_{1e}$). After finetuning, we evaluated the robustness of the embedded watermark by extracting it and measuring the p-value. The results demonstrated a p-value significantly lower than 5e-4, with $N=163840$, indicating the passport $C$ remains detectable.

\subsection{Structural Obfuscation Attack.}
\label{exp:obfuscation_attack}
\begin{figure}[!ht]
    \flushright
    \includegraphics[width=0.85\linewidth]{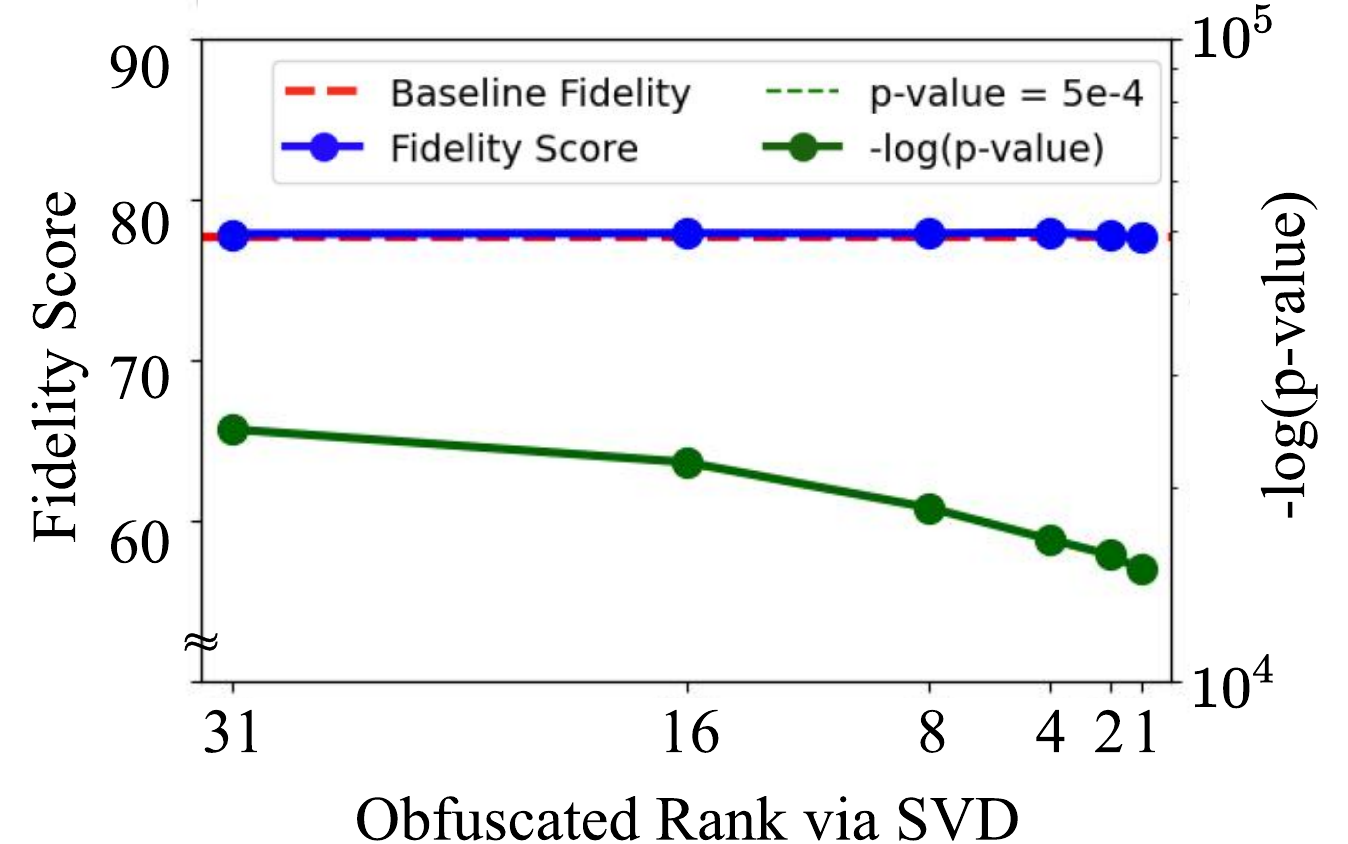}
    \caption{Structural Obfuscation Attack on SEAL weight of Gemma-2B via SVD. The original rank is 32, and the ranks are obfuscated from 31 down to 1.}
    \label{fig:svd_obfuscation}
\end{figure}

Structural obfuscation attacks target the structure of DNN models while maintaining their functionality \cite{yan2023rethinking, pegoraro2024deepeclipse}.
In the case of LoRA, an adversary cannot change the input dimension $a$ or the output dimension $b$, but they can modify the rank $r$ of the matrices $A' \in \mathbb{R}^{r \times a}$ and $B' \in \mathbb{R}^{b \times r}$.
However, even if $r$ is changed, $\mathbb{N}_{obf}$ remains functionally equivalent to $\mathbb{N}$, ensuring the distributed passport $C$ remains detectable.
To mitigate the effects of structural obfuscation with minimal impact on the host task, we decompose $\mathbb{N}(\cdot)$ using SVD and modify it based on its singular values, sorting by large singular values and discarding the smaller ones, resulting in $\mathbb{N} \simeq \mathbb{N}_{svd}$.

\figref{fig:svd_obfuscation} shows the results of performing structural obfuscation via SVD. The original rank is 32, and the results are obfuscated from rank 31 down to 1. The fidelity score remains unchanged, and the passport $C$ is still detectable, demonstrating \scheme{}'s robustness against structural obfuscation attacks.

\subsection{Ambiguity Attack}
\label{sec:robustness_ambiguity}

\begingroup
\setlength{\tabcolsep}{5pt} %
\renewcommand{\arraystretch}{1.0} %

\begin{table}[!ht]
    \centering
    \caption{Fidelity performance, $M_{T}$, table for each passport on commonsense reasoning task, $T$.
}
    \begin{tabular*}{0.79\linewidth}{l|cc|c}
\toprule
  Model &  $C_{t}=C$  & $C_{t}=C_p$ & $\epsilon_{T}$ \\
\midrule
     LLaMA-2-7B  & 82.2 & 82.7 &  0.5\\
    Mistral-7B-v0.1  & 84.2 & 87.9 & 3.7\\
     Gemma-2B    & 76.3 & 76.6 & 0.3\\
\bottomrule
    \end{tabular*}

    \label{tab:passport_perfomance}
\end{table}

\endgroup

In the context of \scheme{}, ambiguity attacks pose a significant threat when an adversary attempts to create counterfeit passports that can bypass the verification process by generating functionally equivalent weights.
Even under the worst-case assumption that the adversary successfully separates $\mathbb{N}(B', A')$ into $\mathbb{N}(\widetilde{B},\widetilde{C},\widetilde{A})$, they must generate another passport, $\widetilde{C}_{p\text{-adv}}$, to form the required quadruplet for the verification by \defref{def:verification}.

\tabref{tab:passport_perfomance} provides the verification thresholds \( \epsilon_T \).
As depicted in~\figref{fig:ambiguity_attack}, the adversary would need to generate a counterfeit passport $\widetilde{C}_{p\text{-adv}}$ that is more than 60\% similar to $C_{p}$ to avoid a significant drop below $\epsilon_T$. Given the concealed nature of $C_{p}$, achieving this level of similarity is practically impossible, which highlights the effectiveness of our approach in maintaining the security of the ownership verification process. In conclusion, \scheme{} significantly reduces the risk of ambiguity attacks by ensuring that counterfeit passports generated without knowledge of $C_{p}$ are unlikely to maintain the required fidelity score.

\section{Conclusion}
We introduced \scheme{}, a novel watermarking scheme specifically tailored for LoRA weights. By inserting a constant matrix \(C\) during LoRA training and factorizing it afterward, our approach enables robust ownership verification without impairing the model’s performance. Empirical results on commonsense reasoning, instruction tuning, and text-to-image tasks confirm both high fidelity and strong resilience against removal, obfuscation, and ambiguity attacks.

Although our experiments focus on LoRA, the core idea—using a non-trainable matrix to entangle trainable parameters—may extend to other parameter-efficient finetuning (PEFT) methods or larger foundation models. Future work will explore generalized forms of this embedding mechanism, aiming to protect a broader range of adaptation techniques while maintaining minimal overhead.

\section*{Impact Statement}
Our scheme helps content creators and organizations safeguard intellectual property in lightweight, easily distributed LoRA-based models. This fosters more open collaboration in AI communities by alleviating concerns about unauthorized use or redistribution of finetuned checkpoints.

However, no defense is fully immune to new adversarial strategies, and watermarking could be misused to embed covert or unethical content. We thus advocate for transparent guidelines and continuous evaluation to ensure that watermarking remains a fair and dependable approach for protecting intellectual property in open-source AI.

\newpage
\appendix
\onecolumn

\clearpage
\onecolumn

\section{Notation}
\label{appendix:notation}
\begingroup
\setlength{\tabcolsep}{4pt} %
\renewcommand{\arraystretch}{1.2} %

\begin{table}[ht]
\centering
\caption{Notation table for \scheme{}. Key symbols and their definitions.}
\label{tab:notation}
\begin{tabular}{>{\centering\arraybackslash}m{0.18\linewidth} >{\raggedright\arraybackslash}m{0.75\linewidth}}
\toprule
\textbf{Symbol} & \textbf{Description} \\
\midrule

$W$ & Pretrained model weight (size $b \times a$) on which LoRA is applied. \\

$B, A$ 
& LoRA’s trainable \emph{up} and \emph{down} blocks, where 
$B \in \mathbb{R}^{b \times r}$, 
$A \in \mathbb{R}^{r \times a}$, 
and $r \ll \min(b,a)$. \\

$B', A'$
& Publicly released LoRA weights \emph{after} distributing the passport $C$ (see Def.\ref{def:decomposition_function}).
These have the same shape as $B, A$. \\

$\Delta W$
& The weight offset from LoRA (or \scheme{}). For instance, 
$\Delta W = B\,C\,A$ or $B\,A$ depending on context. \\

$\mathbb{N}(\cdot)$
& The adaptation layer operator; e.g., $\mathbb{N}(B,A)$ for standard LoRA, or $\mathbb{N}(B,A,C)$ for \scheme{}. \\

$C, C_p$
& Non-trainable \emph{passports} in \scheme{}. 
$C$ is the main passport hidden into $B', A'$; 
$C_p$ is an additional passport for ownership verification. 
Both are in $\mathbb{R}^{r \times r}$. \\

$\widetilde{B}, \widetilde{A}, \widetilde{C}\,(\widetilde{C}_{p\text{-adv}})$
& An \emph{adversarial factorization} of publicly released weights $(B',A')$ that an attacker attempts to construct; e.g.\ $\widetilde{B}\,\widetilde{C}\,\widetilde{A} = B'A'$. 
In some scenarios, an attacker may generate $\widetilde{C}_{p\text{-adv}}$ to forge an additional passport. These have the same shape as $B, A, C$ respectively. \\

$C_t$
& A \textit{runtime passport} (e.g., used in inference or verification) for given $B, A$. \\

$f(\cdot)$
& Decomposition function that takes $C$ and returns two factors $(C_{1}, C_{2})$ such that $C_{1}C_{2} = C$. 
For example, $f_{svd}$ uses Singular Value Decomposition (SVD). \\

$T$
& The \emph{host task} (e.g., instruction following, QA), to which LoRA (\scheme{}) is adapted. \\

$M_{T}(\cdot)$
& A \emph{fidelity score} or performance metric (e.g., accuracy) of the adaptation layer on task $T$. \\

$V(\cdot)$
& The verification process (function) that checks authenticity of passports (Sec.~\ref{sec:verification}). 
It outputs \texttt{True} or \texttt{False}. \\

$\epsilon_T$
& A threshold used in the verification stage to decide ownership claims. \\

\bottomrule
\end{tabular}
\end{table}
\endgroup

\section{Training Process of \scheme{}}
\label{appendix:training_procees_seal}

\subsection{Forward Path}
In \scheme{}, the forward path produces the output \(W'\) by adding a learnable offset \(\Delta W\) on top of the base weights \(W\):
\begin{equation}
\label{eq_seal}
    W' = W + \Delta W = W + BCA.
\end{equation}
Here, \(B\) and \(A\) are trainable matrices, while \(C\) is a fixed \emph{passport} matrix that carries the watermark. Unlike traditional LoRA layers that use \(\Delta W = BA\) alone, \scheme{} inserts \(C\) between \(B\) and \(A\). This additional matrix:
\begin{itemize}
    \item Forces the resulting offset \(\Delta W\) to pass through an extra linear transformation, potentially mixing or reorienting the learned directions.
    \item Ties the final weight update \(\Delta W\) to the presence of \(C\); removing or altering \(C\) would disrupt \(\Delta W\) and hence the model's functionality.
\end{itemize}

\begin{figure}[t]
    \centering
    \includegraphics[width=0.95\linewidth]{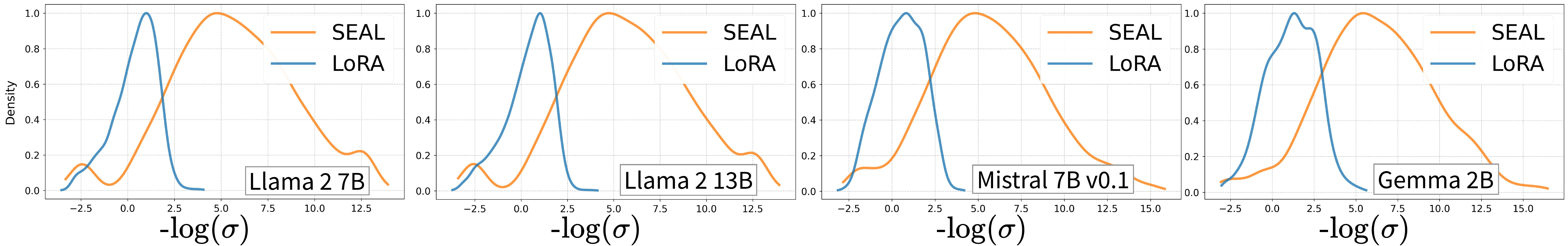}
    \caption{KDE of $-\log(\sigma)$ for LoRA vs.\ SEAL.
    We extract the top-32 singular values $\sigma$ from each module of the finetuned 
    $\Delta W$ (for rank=32 $\mathbb{N}(\cdot)$) and plot $-\log(\sigma)$ via a 
    kernel density estimate (KDE).}
    \label{fig:kde_log_singular_value}
\end{figure}

If \(C\) were diagonal, it would merely scale each dimension independently, which can be easier to isolate or undo. However, when \(C\) is a full (non-diagonal) matrix, the learned offset \(\Delta W\) may exhibit more complex structures, as the multiplication by \(C\) intermixes channels or dimensions. Such a design can lead to a more \emph{wider} singular value distribution (see \figref{fig:kde_log_singular_value}), where the watermark is spread across multiple directions, thus making it less prone to straightforward removal.

\subsection{Backward Path}
The backward path computes gradients of the loss function \(\phi\) with respect to \(A\) and \(B\), revealing how \(C\) influences the updates. Let
\begin{equation}
\label{eq_backward_0}
    \Delta := BCA 
    \quad \text{and} \quad 
    \Phi := \phi(\Delta x),
\end{equation}
where \(\Delta x\) represents applying \(\Delta\) to some input \(x\). Then, by the chain rule,
\begin{align}
\label{eq_backward_A}
    \frac{\partial \Phi}{\partial A} 
    &= (BC)^T \, \frac{\partial \phi}{\partial \Delta} 
     = C^T B^T \, \frac{\partial \phi}{\partial \Delta}, \\[4pt]
\label{eq_backward_B}
    \frac{\partial \Phi}{\partial B} 
    &= \frac{\partial \phi}{\partial \Delta} \, (CA)^T
     = \frac{\partial \phi}{\partial \Delta}\, A^T\,C^T.
\end{align}

These expressions highlight two key points:
\begin{enumerate}
    \item[\textbf{(1)}] \textbf{Transformation of Gradients.}  
        Each gradient, \(\nabla_A\) and \(\nabla_B\), is multiplied (from the left or right) by \(C^T\). If \(C\) were diagonal, this would reduce to element-wise scaling of the gradient, which is relatively simple to reverse or interpret. In contrast, a \emph{full} \(C\) applies a more general linear transformation---potentially a rotation or mixing---to the gradient directions.
    \item[\textbf{(2)}] \textbf{Entanglement of Learnable Parameters.}  
        Because \(C\) is \emph{fixed} but non-trivial, both \(B\) and \(A\) are continually updated in a manner dependent on \(C\). Over many gradient steps, \(\Delta W = BCA\) becomes \emph{entangled} across multiple dimensions; single-direction modifications in \(B\) or \(A\) cannot easily 
        isolate the watermark without affecting other directions. 
\end{enumerate}

\paragraph{Impact on Singular Values.}
This interplay of forward and backward paths explains why \(\Delta W = BCA\) often ends up with a different singular value spectrum than that of a simpler \(\Delta W = BA\). Intuitively, placing \(C\) between \(B\) and \(A\) introduces:
\begin{itemize}
    \item \textbf{Additional mixing in the forward pass}: The matrix product \(B \cdot C \cdot A\) can redistribute any localized pattern in \(B\) or \(A\) across more directions.
    \item \textbf{Gradient reorientation in the backward pass}: The terms $C^T$ in Eqs.~\eqref{eq_backward_A}--\eqref{eq_backward_B} reshape how errors flow back to \(B\) and \(A\), potentially encouraging them to explore a broader subspace. 
\end{itemize}
As a result, the learned update \(\Delta W\) may exhibit a less concentrated singular value distribution, meaning it is not dominated by just a few principal components. Instead, it becomes harder to nullify or compress the watermark without causing broader distortion in the model. 

\paragraph{Practical Advantage.}
Because \(\Delta W\) is effectively ``spread'' across multiple singular directions, any attempt to remove or alter the watermark by targeting a handful of directions is likely to degrade performance. Thus, from both the forward and backward perspectives, \(C\) serves as a robust vehicle for embedding the watermark: 
\begin{enumerate}
    \item It cannot be trivially factored out without retraining \(B\) and \(A\) (forward path).
    \item Its mixing effect on gradients entangles the learned parameters, creating a more diffuse subspace in which the watermark resides (backward path).
\end{enumerate}
These properties collectively bolster \scheme{}'s resistance to watermark removal attacks, while minimally affecting the primary task performance.

\section{On Forging Multiple Passports from a Single Factorization}
\label{appendix:forge_multi_passports}

This section clarifies why an adversary \emph{cannot} simply factorize the released LoRA weights $(B', A')$ into some $(\widetilde{B}, \widetilde{C}, \widetilde{A})$ and then create an additional \textit{passport} $\widetilde{C}_{p\text{-adv}}$ in order to circumvent our multi-passport verification. We also reiterate that \scheme{} is intentionally indistinguishable from a standard LoRA, so an attacker generally cannot even discern that \scheme{} was used.

\subsection{Indistinguishability from Standard LoRA}
\label{appendix:indistinguishability}

By design, the publicly distributed weights are simply $B' \in \mathbb{R}^{b \times r}$ and $A' \in \mathbb{R}^{r \times a}$, analogous to standard LoRA. 
No additional matrix parameters (or suspicious metadata) are visible. 
Hence, without insider knowledge, an attacker cannot tell \emph{a priori} if $(B', A')$ derives from \scheme{} or a conventional LoRA finetuning. 
This alone imposes a significant hurdle:  
\[
  \text{Attacker must first \emph{discover}} \, (\text{or guess}) \, \text{that \scheme{} was used.}
\]
Only then might they attempt forging hidden passports.  

\subsection{Attempting a Single Factorization for Two Passports}
\label{appendix:single_factorization}

Assume, hypothetically, that an attacker somehow knows a given $(B', A')$ came from \scheme{}. 
They might try a factorization of the form:
\[
   (B', A') \quad \longrightarrow \quad (\widetilde{B}, \widetilde{C}, \widetilde{A}),
\]
so that $\widetilde{B}\,\widetilde{C}\,\widetilde{A} = B' A'$. 
Then they could designate $\widetilde{C}$ as a \textit{forged} version of the original $C$. 

\paragraph{Creating a Second Passport.}
Furthermore, to break multi-passport verification (see \secref{sec:verification}), the attacker would need \emph{another} passport, $\widetilde{C}_{p\text{-adv}}$, that also yields near-identical fidelity scores:
\[
   M_T(\mathbb{N}(\widetilde{B}, \widetilde{A}, \widetilde{C})) \;\approx\; 
   M_T(\mathbb{N}(\widetilde{B}, \widetilde{A}, \widetilde{C}_{p\text{-adv}}))
   \quad (\text{for all relevant data for task, }T).
\]
However, this requires that \(\widetilde{B},\widetilde{A}\) be \emph{simultaneously} entangled with \emph{two distinct} passports, which is nontrivial for a single factorization. 

\subsection{Why a Single Factorization Cannot Produce Two Entangled Passports}
\label{appendix:why_no_multi_passports}

\begin{itemize}
    \item \textbf{Concurrent Entanglement is Required.} 
    In \scheme{}, $B$ and $A$ are co-trained (entangled) with both $C$ and $C_{p}$ \emph{at the same time} during finetuning. 
    This ensures that, for any batch, either $C$ or $C_p$ is used, such that $B,A$ adapt to \emph{both} passports. 
    Merely performing a \emph{post-hoc} factorization on $(B', A')$ does not replicate this simultaneous learning process.

    \item \textbf{One Factorization Yields One Mapping.} 
    A single factorization typically captures \emph{one} equivalence, e.g.\ $\widetilde{C}$. 
    Generating an additional $\widetilde{C}_{p\text{-adv}}$ that \emph{also} achieves the same function (or fidelity) using the \emph{same} $\widetilde{B}, \widetilde{A}$ is a significantly more constrained problem. 
    In practice, an attacker would need to re-finetune $(\widetilde{B}, \widetilde{A})$ \emph{twice}, once for each passport, effectively mimicking the original training—but without knowledge of the original dataset $\mathcal{D}$.

    \item \textbf{Costly and Uncertain Outcome.} 
    Even if the attacker invests major computational resources, re-training two passports from scratch is as expensive as (or more expensive than) training a brand-new LoRA model. 
    Moreover, success is not guaranteed, since the attacker must ensure $\widetilde{C}_{p\text{-adv}} \neq \widetilde{C}$ but still replicates near-identical behavior on the entire dataset, all while not knowing the original dataset $\mathcal{D}$ or training schedule.
\end{itemize}

\subsection{Proof of Non-Existence of Two Distinct Passports from One Factorization}
\label{sec:proof_one_factorization}

\paragraph{Assumptions.}
We assume the attacker fixes rank-$r$ matrices
\[
    \widetilde{B} \in \mathbb{R}^{b \times r}, 
    \quad
    \widetilde{A} \in \mathbb{R}^{r \times a}
    \quad\text{with}\quad
    \mathrm{rank}(\widetilde{B}) = \mathrm{rank}(\widetilde{A}) = r.
\]
This aligns with standard LoRA dimensionality and preserves maximum utility (see Remark~\ref{remark:why_rank_r} below).

\paragraph{Statement.}
Suppose the attacker finds two different passports, $\widetilde{C} \neq \widetilde{C}_{p\text{-adv}}$, each in $\mathbb{R}^{r\times r}$, satisfying
\[
    \widetilde{B}\,\widetilde{C}\,\widetilde{A} \;=\; 
    \widetilde{B}\,\widetilde{C}_{p\text{-adv}}\,\widetilde{A} 
    \;=\;
    B'A'.
\]
We show this leads to a contradiction.

\paragraph{Pseudo-inverse argument (short version).}
If the attacker specifically uses the pseudoinverse-based approach,
\[
    \widetilde{C} 
    \;=\; 
    \widetilde{B}^\dagger \,(B'A')\,\widetilde{A}^\dagger,
    \quad
    \widetilde{C}_{p\text{-adv}}
    \;=\; 
    \widetilde{B}^\dagger \,(B'A')\,\widetilde{A}^\dagger,
\]
then clearly $\widetilde{C} = \widetilde{C}_{p\text{-adv}}$, contradicting $\widetilde{C}\neq \widetilde{C}_{p\text{-adv}}$.

\paragraph{More general linear algebra argument (rank-$r$).}
Even without explicitly constructing $\widetilde{B}^\dagger$ or $\widetilde{A}^\dagger$, one can show:
\[
    \bigl(\widetilde{C} - \widetilde{C}_{p\text{-adv}}\bigr)
    \;\;\Longrightarrow\;\;
    \widetilde{B}\,\bigl(\widetilde{C} - \widetilde{C}_{p\text{-adv}}\bigr)\,\widetilde{A}
    \;=\; O.
\]
Since $\widetilde{B},\widetilde{A}$ each have rank $r$, this forces $\widetilde{C} - \widetilde{C}_{p\text{-adv}} = O$, implying $\widetilde{C}=\widetilde{C}_{p\text{-adv}}$. Hence, no two distinct passports can arise from the same factorization $(\widetilde{B}, \widetilde{A})$.

\begin{equation}
\label{eq:adv_factorization_contradiction}
\widetilde{C} \;\neq\; \widetilde{C}_{p\text{-adv}}
\quad\Longrightarrow\quad
\widetilde{C} - \widetilde{C}_{p\text{-adv}} \;=\; O.
\quad\text{(Contradiction)}
\end{equation}

\paragraph{Remark on rank-deficient factorizations.}
\label{remark:why_rank_r}
If $\widetilde{B}$ or $\widetilde{A}$ has rank $< r$, then infinitely many $\widetilde{C}$ can satisfy $\widetilde{B}\,\widetilde{C}\,\widetilde{A} = B'A'$. However, such rank-deficient choices almost always degrade the model’s fidelity (losing degrees of freedom), thus failing to preserve the same performance as $(B', A')$. Consequently, attackers seeking to maintain \emph{full utility} have no incentive to choose rank-deficient $\widetilde{B}, \widetilde{A}$. Therefore, we assume $\mathrm{rank}(\widetilde{B})=\mathrm{rank}(\widetilde{A})=r$ to ensure that $(B'A')$ is matched faithfully.

\subsection{No Practical Payoff for Such an Attack}
\begin{enumerate}
    \item \textbf{Attackers Typically Lack Data.} 
    To even begin constructing $(\widetilde{C}, \widetilde{C}_{p\text{-adv}})$, attackers must have \textit{access} to the original training data (or certain proportion of dataset with similar distribution) \emph{and} be certain \scheme{} was used. 
    Both are high barriers. Training dataset is not a part of \scheme{}, and is mostly proprietary. It does not violate Kerckhoff's principal. 

    \item \textbf{Equivalent to Costly Re-Training.} 
    Producing two passports that match all fidelity checks essentially replicates the original multi-passport entanglement from scratch. 
    This yields no distinct advantage over simply training a new LoRA.

    \item \textbf{Cannot Disprove Legitimate Ownership.} 
    Even if they succeed in forging $\widetilde{C}, \widetilde{C}_{p\text{-adv}}$, the legitimate owner’s original pair $(C, C_p)$ still correctly verifies, preserving the rightful ownership claim.
\end{enumerate}

\subsection{Conclusion}

In summary, forging multiple passports from a single factorization of $(B', A')$ is infeasible because \scheme{}’s multi-passport structure relies on \emph{concurrent} entanglement of $B,A$ with \emph{both} passports $C$ and $C_p$ during training.  
A single post-hoc factorization can at best replicate \emph{one} equivalent mapping, but not \emph{two} functionally interchangeable mappings without a re-finetuning process that is as expensive and uncertain as building a new model.  
Furthermore, since \scheme{} weights are indistinguishable from standard LoRA, the attacker generally cannot even detect the scheme in the first place. 
Therefore, this approach does not offer a viable pathway to break or circumvent \scheme{}’s multi-passport verification procedure.

\section{Extensions to Matmul-based LoRA Variants}
\label{appendix:matmul_variants}

Beyond the canonical LoRA~\cite{hu2021lora} formulation, numerous follow-up works propose modifications and enhancements while still employing matrix multiplication (\texttt{matmul}) as the underlying low-rank adaptation operator. In this section, we illustrate how \scheme{} is compatible or can be adapted to these matmul-based variants. Although we do not exhaustively enumerate every LoRA-derived approach, the general principle remains: if the adaptation primarily uses matrix multiplication (possibly with additional diagonal, scaling, or regularization terms), then \scheme{} can often be inserted by embedding a non-trainable passport $C$ between the \emph{up} and \emph{down} blocks.%

\subsection{LoRA-FA~\cite{zhang2023lora}}
\label{appendix:matmul_variants_lora_fa}
LoRA-FA (LoRA with frozen down blocks) modifies LoRA by keeping the \emph{down} block frozen during training, while only the \emph{up} block is trained. Structurally, however, it does not alter the fundamental $\texttt{matmul}$ operator. Consequently, integrating \scheme{} follows the same procedure as standard LoRA: one can embed the passport $C$ into the product $B\,C\,A$ without requiring any special adjustments. The difference in training rules (i.e.\ freezing $A$) does not affect how $C$ is placed or how it is decomposed into $(C_1, C_2)$ for final public release.

\subsection{LoRA+~\cite{hayoulora+}}
\label{appendix:matmul_variants_lora_plus}
LoRA+ investigates the training dynamics of LoRA’s \emph{up} ($B$) and \emph{down} ($A$) blocks. In particular, it emphasizes the disparity in gradient magnitudes and proposes using different learning rates:
\[
    A \;\leftarrow\; A \;-\; \eta \; G_{A}, 
    \quad
    B \;\leftarrow\; B \;-\; \lambda \,\eta \; G_{B},
\]
where $\lambda \gg 1$ is a scale factor, $\eta$ is the base learning rate, and $G_{A}, G_{B}$ are the respective gradients. LoRA+ does \emph{not} alter the structural operator (still matrix multiplication). Therefore, \scheme{} can be employed by introducing $C \in \mathbb{R}^{r \times r}$ between $B$ and $A$, yielding $\Delta W = B\,C\,A$. The difference in gradient scaling does not impact the usage of a non-trainable passport matrix $C$.

\subsection{VeRA~\cite{kopiczkovera}}
\label{appendix:matmul_variants_vera}
VeRA introduces two diagonal matrices, $\Lambda_b$ and $\Lambda_d$, to scale different parts of the low-rank factors:
\[
    \Delta W \;=\; \Lambda_{b} \, B \, \Lambda_{d} \, A,
\]
where $B, A$ may be random, frozen, shared across layers and the diagonal elements in $\Lambda_b, \Lambda_d$ are trainable. Despite these diagonal scalings, the core operator remains matrix multiplication. Hence, embedding a passport $C$ is still feasible. By leveraging the commutative property of diagonal matrices and $C$ (assuming $C$ commutes with $\Lambda_d$ in the sense that one can re-factor $C$ into $C_1 \Lambda_d C_2$ or $\Lambda_d C$), \scheme{} can be inserted:

\[
    \Delta W \;=\; \Lambda_b \,(B\,C_1)\, \Lambda_d \,(C_2 \,A),
\]
which is functionally identical to $\Lambda_b \,B\,\Lambda_d \,A$ except for the hidden passport $C = C_1 C_2$. Implementing \scheme{} in VeRA may require converting the final trained weights back into a standard $(B', A')$ form plus a diagonal scaling term, but the fundamental principle is straightforward.

\subsection{AdaLoRA~\cite{zhangadaptive}}
\label{appendix:matmul_variants_adalora}
AdaLoRA applies a dynamic rank-allocating approach inspired by SVD. It factorizes the weight update into:
\[
    \Delta W \;=\; P \,\Lambda\, Q,
\]
where $\Lambda$ is a diagonal matrix, and $P, Q$ are regularized to maintain near-orthogonality. Since diagonal matrices commute under multiplication (up to a re-factorization), one can embed a passport $C$ by decomposing it ($f(C) \rightarrow (C_1, C_2)$). In essence,
\[
    \Delta W \;=\; P\,C_1 \,\Lambda\, C_2\, Q 
    \;\;=\;\; P' \,\Lambda\, Q',
\]
where $P' = PC_1$ and $Q' = C_2 Q$. This preserves the rank-$r$ structure and does not disrupt AdaLoRA’s optimization logic. Regularization terms that enforce $P'^T P' \approx I$ and $Q'Q'^T \approx I$ remain valid, though one may incorporate $C_1, C_2$ into the initialization or adapt them carefully so as not to degrade the orthogonality constraints.

\subsection{DoRA~\cite{liu2024dora}}
\label{appendix:matmul_variants_dora}
DoRA modifies the final LoRA update using a column-wise norm factor:
\[
    W' 
    \;=\;
    \frac{\|W\|_c}{\|\,W +\Delta W\,\|_c}\,\bigl(W + \Delta W\bigr),
\]
where $\| \cdot \|_c$ computes column-wise norms and the ratio is (by design) often detached from gradients to reduce memory overhead. Replacing $\Delta W$ with $B\,C\,A$ in DoRA does not alter the external gradient manipulation logic, since $C$ is non-trainable. Thus,
\[
    W' 
    \;=\;
    \frac{\|W\|_c}{\|\,W + B\,C\,A\,\|_c}\,\bigl(W + B\,C\,A\bigr)
\]
remains valid. The presence of $C$ does not interfere with DoRA’s approach to scaling or norm-based constraints.

\subsection{Variants with Non-Multiplicative Operations}
\label{appendix:matmul_variants_nonmult}
All of the above variants preserve the core LoRA assumption of a matrix multiplication operator for the rank-$r$ adaptation. However, certain approaches introduce non-multiplicative adaptations (e.g., Hadamard product, Kronecker product, or other specialized transforms). In the following section, for these cases, which discuss how \scheme{} can be generalized to any bilinear or multilinear operator $\star$.

\section{Extensions to Generalized Low-Rank Operators}
\label{appendix:generalized_operators}

In the main text, we considered a standard LoRA \cite{hu2021lora} that uses a matrix multiplication operator:
\[
    \Delta W = B \; C \; A,
\]
where $B \in \mathbb{R}^{b \times r}$, $C \in \mathbb{R}^{r \times r}$, and $A \in \mathbb{R}^{r \times a}$. 
Recent work has explored alternative low-rank adaptation mechanisms beyond simple \texttt{matmul}, such as Kronecker product-based methods~\cite{edalati2022krona,yeh2023navigating} or even elementwise (Hadamard) product~\cite{hyeon2021fedpara} forms. 
Our approach can be extended in a straightforward manner to these generalized operators, which we denote as $\star$. 

\subsection{General Operator \texorpdfstring{$\star$}{*}}
Let $\star$ be any bilinear or multilinear operator used for low-rank adaptation.\footnote{Here, \emph{bilinear} means $(X \star Y)$ is linear in both $X$ and $Y$ when one is held fixed, e.g.\ standard matrix multiplication, Kronecker product, or Hadamard product.}
We can then write the trainable adaptation layer as
\[
    \Delta W = B \;\star\; C \;\star\; A,
\]
where $B, A$ are the trainable low-rank parameters, and $C$ is the non-trainable passport in \scheme{}. 
During training, $B$ and $A$ are optimized in conjunction with $C$ held fixed (just as in the matrix multiplication case).

\paragraph{Decomposition Function for Operator \texorpdfstring{$\star$}{*}.}
To \emph{distribute} $C$ into $(B, A)$ after training, we require a \emph{decomposition function} $f: C \mapsto (C_1, C_2)$ such that
\[
    C = C_1 \;\star\; C_2.
\]
For example, under the Kronecker product $\otimes$, one could define $f(C)$ to split $C$ into smaller block partitions, or use an SVD-like factorization in an appropriate transformed space. Under the Hadamard product, $f(C)$ could involve elementwise roots or other transformations.

Once $C_1$ and $C_2$ are obtained, we apply:
\[
    B' \;=\; B \;\star\; C_1
    \quad,\quad
    A' \;=\; C_2 \;\star\; A,
\]
so that
\[
    B' \;\star\; A' 
    \;=\; 
    (B \;\star\; C_1) \star (C_2 \star A)
    \;=\; B \;\star\; (C_1 \star C_2) \;\star\; A 
    \;=\; 
    B \;\star\; C \;\star\; A.
\]
Hence, the final distributed weights $(B', A')$ for public remain \emph{functionally equivalent} to using $B, A, C$.

\subsection{Implications and Future Directions}

\begin{itemize}
    \item \textbf{Broader Applicability.} 
    By permitting $\star$ to be any bilinear or multilinear operator (Kronecker, Hadamard, etc.), \scheme{} naturally extends beyond the canonical matrix multiplication used in most LoRA implementations. 
    This flexibility can be valuable for advanced parameter-efficient tuning methods \cite{edalati2022krona,hyeon2021fedpara,yeh2023navigating}.

    \item \textbf{Same Security Guarantees.}
    The central watermarking principle (embedding a non-trainable passport $C$ into the adaptation) does not change. 
    An adversary attempting to re-factor $B' \star A'$ to recover $C$ faces the same challenges described in the main text and \suppl{} \ref{appendix:forge_multi_passports}—non-identifiability, cost of reconstruction, and multi-passport verification barriers.

    \item \textbf{Potential Operator-Specific Designs.}
    Certain operators (e.g., Kronecker product) may admit additional constraints or factorization strategies that could be exploited for improved stealth or efficiency. 
    Investigating these is an interesting direction for future work. 
\end{itemize}

\noindent
In summary, \scheme{} can be generalized to other operators $\star$ by treating $C$ as a non-trainable factor and defining a suitable decomposition function $f(C)$ such that $C = C_1 \,\star\, C_2$. 
This allows us to hide the passport just as in the matrix multiplication case, thereby preserving the main \scheme{} pipeline for more complex LoRA variants.

\section{Training Details}
\label{appendix:training_details}

\subsection{Commonsense Reasoning Tasks}
\begin{table*}[t]
    \caption{Hyperparameter configurations of SEAL and LoRA for Gemma-2B, Mistral-7B-v0.1, LLaMA2-7B/13B, and LLaMA3-8B on the commonsense reasoning. All experiments are done with 4x A100 80GB (for LLaMA-2-13B) and 4x RTX 3090 (for the other models) with approximately 15 hours.}

    \centering
    \setlength{\tabcolsep}{4pt} %
    \begin{tabular}{lcccccccccc}
        \toprule
        Models & \multicolumn{2}{c}{Gemma-2B} & \multicolumn{2}{c}{Mistral-7B-v0.1} & \multicolumn{2}{c}{LLaMA-2-7B} & \multicolumn{2}{c}{LLaMA-2-13B} & \multicolumn{2}{c}{LLaMA-3-8B} \\
        \midrule
        Method & LoRA & SEAL & LoRA & SEAL & LoRA & SEAL & LoRA & SEAL & LoRA & SEAL \\
        r & \multicolumn{10}{c}{32} \\
        alpha & \multicolumn{10}{c}{32} \\
        Dropout & \multicolumn{10}{c}{0.05} \\
        LR & 2e-4 & 2e-5 & 2e-5 & 2e-5 & 2e-4 & 2e-5 & 2e-4 & 2e-5 & 2e-4 & 2e-5 \\
        Optimizer & \multicolumn{10}{c}{AdamW~\cite{loshchilovdecoupled}} \\
        LR scheduler & \multicolumn{10}{c}{Linear} \\
        Weight Decay & \multicolumn{10}{c}{0} \\
        Warmup Steps & \multicolumn{10}{c}{100}  \\
        Total Batch size & \multicolumn{10}{c}{16} \\
        Epoch & \multicolumn{10}{c}{3} \\
        Target Modules & \multicolumn{10}{c}{Query Key Value UpProj DownProj} \\
        \bottomrule
    \end{tabular}
    \label{tab:hparams_commonsense_reasoning}
\end{table*}

We conduct evaluations on commonsense reasoning tasks using eight distinct sub-tasks: Boolean Questions (\textbf{BoolQ}) \cite{clark2019boolq}, Physical Interaction QA (\textbf{PIQA}) \cite{bisk2020piqa}, Social Interaction QA (\textbf{SIQA}) \cite{sap2019socialiqa}, Narrative Completion (\textbf{HellaSwag}) \cite{zellers2019hellaswag}, Winograd Schema Challenge (\textbf{Wino}) \cite{sakaguchi2021winogrande}, ARC Easy (\textbf{ARC-e}), ARC Challenge (\textbf{ARC-c}) \cite{clark2018think}, and Open Book QA (\textbf{OBQA}) \cite{OpenBookQA2018}.

We benchmark \scheme{} and LoRA against LLaMA-2-7B/13B \cite{touvron2023llama}, LLaMA-3-8B \cite{llama3modelcard}, Gemma-2B \cite{team2024gemma}, and Mistral-7B-v0.1 \cite{jiang2023mistral} across these commonsense reasoning tasks.

The hyperparameters used for these evaluations are listed in  \tabref{tab:hparams_commonsense_reasoning}.

\subsection{Textual Instruction Tuning}

\begingroup
\setlength{\tabcolsep}{1.6pt} %
\renewcommand{\arraystretch}{1} %

\begin{table*}[t]
    \caption{Hyperparameter configurations of SEAL and LoRA for Instruction Tuning. All experiments are done with 1x A100 80GB for approximately 2 hours. All w/o LM HEAD are Query, Key, Value, Out, UpProj, DownProj, GateProj.}

    \centering
    \begin{tabular}{lcc}
        \toprule
        Model & \multicolumn{2}{c}{LLaMA-2-7B} \\
        \midrule
        Method & LoRA & SEAL \\
        r & \multicolumn{2}{c}{32} \\
        alpha & \multicolumn{2}{c}{32} \\
        Dropout & \multicolumn{2}{c}{0.0} \\
        LR & \multicolumn{2}{c}{2e-5} \\
        LR scheduler & \multicolumn{2}{c}{Cosine} \\
        Optimizer & \multicolumn{2}{c}{AdamW} \\
        Weight Decay & \multicolumn{2}{c}{0} \\
        Total Batch size & \multicolumn{2}{c}{8} \\
        Epoch & \multicolumn{2}{c}{3} \\
        Target Modules & \multicolumn{2}{c}{All w/o LM HEAD} \\
        \bottomrule
    \end{tabular}
    \label{tab:inst_tune_hyperparameters}
\end{table*}

\endgroup

We conducted textual instruction tuning using Alpaca dataset \cite{alpaca} on LLaMA-2-7B \cite{touvron2023llama}, trained for 3 epochs. The hyperparameters used for this process are detailed in \tabref{tab:inst_tune_hyperparameters}.

\subsection{Viusal Instruction Tuning}

\begin{table*}[t]
    \caption{Performance comparison of different methods across seven visual instruction tuning benchmarks}
    \centering
    \begin{tabular}{lccccccccc}
        \toprule
        Method & \# Params (\%) & VQAv2 & GQA & VisWiz & SQA & VQAT & POPE & MMBench & Avg \\
        \midrule
        FT & 100 & 78.5 & 61.9 & 50.0 & 66.8 & 58.2 & 85.9 & 64.3 & 66.5 \\
        LoRA & 4.61 & 79.1 & 62.9 & 47.8 & 68.4 & 58.2 & 86.4 & 66.1 & \textbf{66.9} \\
        SEAL & 4.61 & 75.4 & 58.3 & 41.6 & 66.9 & 52.9 & 86.0 & 60.5 & 63.1\\
        \bottomrule
    \end{tabular}
    \label{tab:vision_instruction_tuning_detailed}
\end{table*}

\begin{table}[!h]
  \caption{Hyperparameters for visual instruction tuning. All experiments were performed with 4x A100 80GB with approximately 24 hours}

  \centering
    \begin{tabular}{lcc}
        \toprule
        Model & \multicolumn{2}{c}{LLaVA-1.5-7B} \\
        \midrule
        Method & LoRA & SEAL\\
        r & \multicolumn{2}{c}{128} \\
        alpha & \multicolumn{2}{c}{128} \\
        LR & 2e-4 & 2e-5\\
        LR scheduler & \multicolumn{2}{c}{Linear} \\
        Optimizer & \multicolumn{2}{c}{AdamW} \\
        Weight Decay & \multicolumn{2}{c}{0} \\
        Warmup Ratio & \multicolumn{2}{c}{0.03}  \\
        Total Batch size & \multicolumn{2}{c}{64} \\
        \bottomrule
    \end{tabular}
  \label{tab:visit_hyperparameters}
\end{table}

\label{appendix:visit_results}
We compared the fidelity of SEAL, LoRA, and FT on the visual instruction tuning tasks with LLaVA-1.5-7B \cite{liu2024llava}.
To ensure a fair comparison, we used the same original model provided by \cite{liu2024llava} uses the same configuration as the LoRA setup with the same training dataset.
We adhere to \cite{liu2024llava} setting to filter the training data and design the tuning prompt format.
 The finetuned models are subsequently assessed on seven vision-language benchmarks: VQAv2\cite{goyal2017vqav2}, GQA\cite{hudson2019gqa}, VisWiz\cite{gurari2018vizwiz}, SQA\cite{lu2022sqa}, VQAT\cite{singh2019vqat}, POPE\cite{li2023pope}, and MMBench\cite{liu2023mmbench}.

\subsection{Text-to-Image Synthesis}

\label{appendix:t2i}
\begingroup
\setlength{\tabcolsep}{8pt}
\renewcommand{\arraystretch}{1.2}
\vspace{0.5cm}
\begin{table}[h!]
    \centering
    \caption{DreamBooth text prompts used for evaluation of inanimate objects and live subjects.}
    \label{tab:dreambooth_prompts}
    \begin{tabular}{|>{\raggedright\arraybackslash}p{0.45\textwidth}|>{\raggedright\arraybackslash}p{0.45\textwidth}|}
    \hline
    \textbf{Prompts for Non-Live Objects} & \textbf{Prompts for Live Subjects} \\ \hline
    a \{\} in the jungle & a \{\} in the jungle \\ \hline
    a \{\} in the snow & a \{\} in the snow \\ \hline
    a \{\} on the beach & a \{\} on the beach \\ \hline
    a \{\} on a cobblestone street & a \{\} on a cobblestone street \\ \hline
    a \{\} on top of pink fabric & a \{\} on top of pink fabric \\ \hline
    a \{\} on top of a wooden floor & a \{\} on top of a wooden floor \\ \hline
    a \{\} with a city in the background & a \{\} with a city in the background \\ \hline
    a \{\} with a mountain in the background & a \{\} with a mountain in the background \\ \hline
    a \{\} with a blue house in the background & a \{\} with a blue house in the background \\ \hline
    a \{\} on top of a purple rug in a forest & a \{\} on top of a purple rug in a forest \\ \hline
    a \{\} with a wheat field in the background & a \{\} wearing a red hat \\ \hline
    a \{\} with a tree and autumn leaves in the background & a \{\} wearing a santa hat \\ \hline
    a \{\} with the Eiffel Tower in the background & a \{\} wearing a rainbow scarf \\ \hline
    a \{\} floating on top of water & a \{\} wearing a black top hat and a monocle \\ \hline
    a \{\} floating in an ocean of milk & a \{\} in a chef outfit \\ \hline
    a \{\} on top of green grass with sunflowers around it & a \{\} in a firefighter outfit \\ \hline
    a \{\} on top of a mirror & a \{\} in a police outfit \\ \hline
    a \{\} on top of the sidewalk in a crowded street & a \{\} wearing pink glasses \\ \hline
    a \{\} on top of a dirt road & a \{\} wearing a yellow shirt \\ \hline
    a \{\} on top of a white rug & a \{\} in a purple wizard outfit \\ \hline
    a red \{\} & a red \{\} \\ \hline
    a purple \{\} & a purple \{\} \\ \hline
    a shiny \{\} & a shiny \{\} \\ \hline
    a wet \{\} & a wet \{\} \\ \hline
    a cube shaped \{\} & a cube shaped \{\} \\ \hline
    \end{tabular}
\end{table}
\endgroup

The DreamBooth dataset~\cite{ruiz2023dreambooth} encompasses 30 distinct subjects from 15 different classes, featuring a diverse array of unique objects and live subjects, including items such as backpacks and vases, as well as pets like cats and dogs. Each of the subjects contains 4-6 images. These subjects are categorized into two primary groups: inanimate objects and live subjects/pets. Of the 30 subjects, 21 are dedicated to objects, while the remaining 9 represent live subjects/pets.

For subject fidelity, following \cite{ruiz2023dreambooth}, we use CLIP-I, DINO. CLIP-I, an image-text similarity metric, compares the CLIP \cite{pmlr-v139-radford21a} visual features of the generated images with those of the same subject images. DINO \cite{caron2021emerging}, trained in a self-supervised manner to distinguish different images, is suitable for comparing the visual attributes of the same object generated by models trained with different methods. For prompt fidelity, the image-text similarity metric CLIP-T compares the CLIP features of the generated images and the corresponding text prompts without placeholders, as mentioned in \cite{ruiz2023dreambooth, nam2024dreammatcher}. For the evaluation, we generated four images for each of the 30 subjects and 25 prompts, resulting in a total of 3,000 images. The prompts used for this evaluation are identical to those originally used in \cite{ruiz2023dreambooth} to ensure consistency and comparability across models. These prompts are designed to evaluate subject fidelity and prompt fidelity across diverse scenarios, as detailed in \tabref{tab:dreambooth_prompts}

\begin{figure*}[!ht]   
    \caption{Comparison of LoRA and \scheme{} in Text-to-Image Synthesis}
    \centering
    \includegraphics[width=\columnwidth]{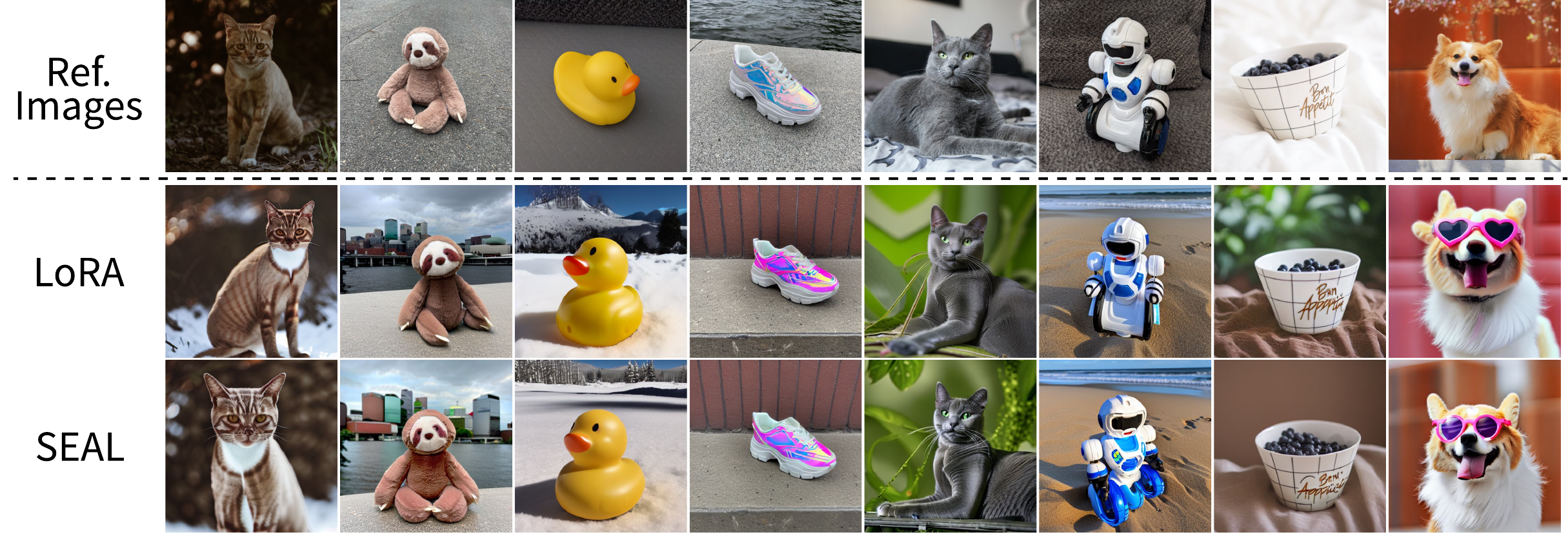}
    \label{fig:t2i_results}
\end{figure*}

\figref{fig:t2i_results} visually compares LoRA and \scheme{} 
on representative subjects from the DreamBooth dataset. 
The top row shows example reference images for each subject, 
the middle row shows images generated by LoRA, 
and the bottom row shows images from our \scheme{}. 
Qualitatively, both methods faithfully capture key attributes of each subject 
(e.g., shape, color, general pose) and produce images of comparable visual quality. 
That is, \scheme{} does not degrade or alter the original subject’s appearance 
relative to LoRA, suggesting that incorporating the constant matrix $C$ 
does not introduce noticeable artifacts or reduce fidelity. 
These results align with the quantitative metrics on subject and prompt fidelity, 
indicating that \scheme{} maintains a quality level on par with LoRA 
while embedding a watermark in the learned parameters.

\begin{table}[!ht]
    \caption{Hyperparameter configurations of SEAL and LoRA for Text-to-Image Synthesis. All experiments are done with 4x RTX 4090 with approximately 15 minutes per subject.}

    \centering
    \begin{tabular}{l>{\centering}p{2cm}>{\centering\arraybackslash}p{2cm}}
        \toprule
        Model & \multicolumn{2}{c}{Stable Diffusion 1.5} \\
        \midrule
        Method & LoRA & SEAL \\
        r & \multicolumn{2}{c}{32} \\
        alpha & \multicolumn{2}{c}{32} \\
        Dropout & \multicolumn{2}{c}{0.0} \\
        LR & 5e-5 & 1e-5 \\
        LR scheduler & \multicolumn{2}{c}{Constant} \\
        Optimizer & \multicolumn{2}{c}{AdamW} \\
        Weight Decay & \multicolumn{2}{c}{1e-2} \\
        Total Batch size & \multicolumn{2}{c}{32} \\
        Steps & \multicolumn{2}{c}{300} \\
        Target Modules & \multicolumn{2}{c}{Q K V Out AddK AddV} \\
        \bottomrule
    \end{tabular}
    \label{tab:t2i_hyperparameters}
\end{table}

\begin{table*}[!ht]
    \centering
    \caption{Hyperparameter configurations of Finetruning Attack on SEAL which trains on 3-epoch. We resume training on $\mathbb{N}(B', A')$, which passport $C$ is distributed in $B, A$ via $f_{svd}$.}

    \begin{tabular}{lccc}
        \toprule
        Model &  LLaMA-2-7B  \\
        \midrule
        Method & LoRA  \\
        r & 32 \\
        alpha & 32 \\
        LR & 2e-5 \\
        Optimizer & AdamW \\
        LR scheduler & Linear \\
        Weight Decay & 0 \\
        Warmup Steps & 100 \\
        Batch size & 16 \\
        Epoch & 1 \\
        Target Modules & Query Key Value UpProj DownProj \\
        \bottomrule
    \end{tabular}
    \label{tab:hparams_finetuning_attack}
\end{table*}

\begin{table*}[t]
    \caption{Hyperparameter configurations of Integrating with DoRA.}
    \centering
    \setlength{\tabcolsep}{4pt} %
    \begin{tabular}{lcccc}
        \toprule
        Model &  \multicolumn{4}{c}{LLaMA-2-7B}  \\
        \midrule
        Method & LoRA & SEAL & DoRA & SEAL+DoRA \\
        r & \multicolumn{4}{c}{32} \\
        alpha & \multicolumn{4}{c}{32} \\
        Dropout & \multicolumn{4}{c}{0.05} \\
        LR & 2e-4 & 2e-5 & 2e-4 & 2e-5  \\
        Optimizer & \multicolumn{4}{c}{AdamW} \\
        LR scheduler & \multicolumn{4}{c}{Linear} \\
        Weight Decay & \multicolumn{4}{c}{0} \\
        Warmup Steps & \multicolumn{4}{c}{100}  \\
        Total Batch size & \multicolumn{4}{c}{16} \\
        Epoch & \multicolumn{4}{c}{3} \\
        Target Modules & \multicolumn{4}{c}{Query Key Value UpProj DownProj} \\
        \bottomrule
    \end{tabular}
    \label{tab:hparams_commonsense_reasoning}
\end{table*}

\clearpage

\section{Ablation Study}

\subsection{Passport Example}
\label{subsec:passport_example}

\begin{figure}[t]
    \centering
    \includegraphics[width=0.4\linewidth]{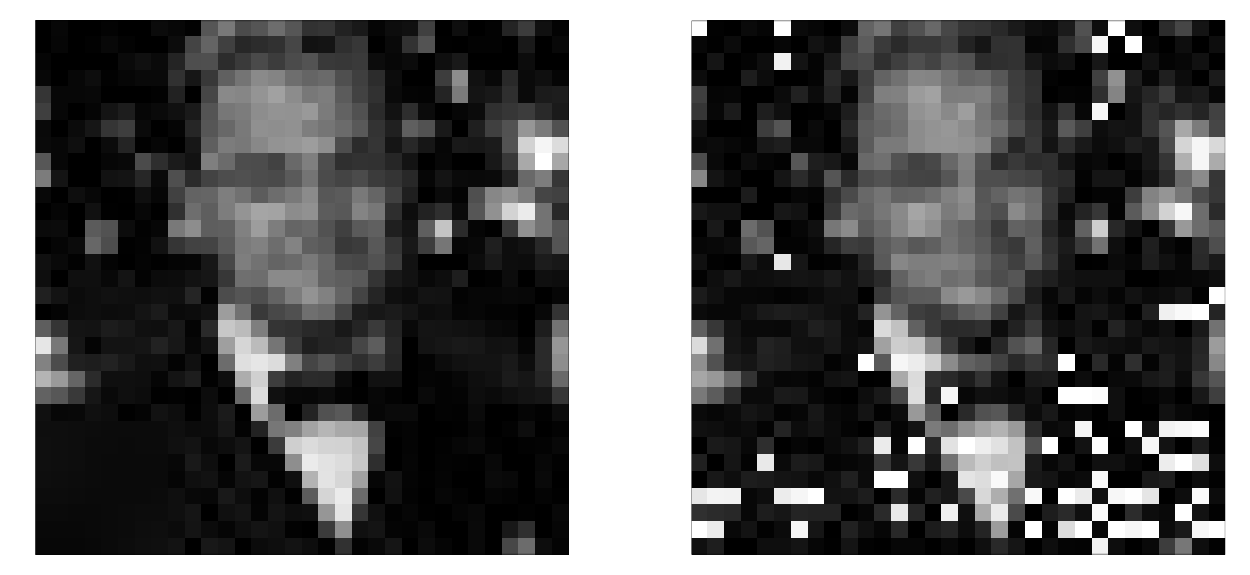}
    \caption{Passport Example.
    \emph{Left}: A 32$\times$32 grayscale bitmap (cropped and downsampled 
    from a YouTube clip\protect\footnotemark) 
    serves as our non-trainable passport $C$.
    \emph{Right}: The passport partially recovered (from 10\% zeroed \scheme{} weight on LLaMA-2-7B).}
    \label{fig:passport_example}
\end{figure}

In order to provide a concrete illustration of our watermark extraction process,
we construct a small 32$\times$32 grayscale image as the \emph{passport} $C$ (or $C_p$).
Specifically, we sampled 100 frames from a publicly available YouTube clip,
applied center-cropping on each frame, converted them to grayscale, 
and then downsampled to 32$\times$32. 
From these frames, we selected one representative image (shown in~\figref{fig:passport_example})
to embed as the non-trainable matrix $C$ in our \scheme{} pipeline~\secref{sec:entangle_passport}.

This tiny passport image, while derived from a movie clip, is both 
\emph{unrecognizable at 32$\times$32} and used exclusively for educational, non-commercial purposes. 
Nevertheless, it visually demonstrates how a low-resolution bitmap 
can be incorporated into the model’s parameter space 
and later \emph{extracted} (possibly with minor distortions) 
to verify ownership.

\footnotetext{\url{https://www.youtube.com/watch?v=2zHHkSu1br4}}

\subsection{Rank Ablation}

To evaluate versatility of the proposed \scheme{} method under varying configurations, we conducted additional experiments focusing on different rank settings (4, 8, 16). The results are summarized in \tabref{tab:rank_ablation_study}. We used the Gemma-2B model \cite{team2024gemma} on commonsense reasoning tasks, as described previously. For comparison, we included the results of LoRA with $r=32$ and \scheme{} with $r=32$ as mentioned in \tabref{tab:it_vit_t2i_all_in_one}.

\begin{table*}[h]
\centering
\caption{Accuracy across various rank settings on commonsense reasoning tasks. The table includes results for rank configurations (4, 8, 16) of \scheme{}, as well as LoRA r=32 and SEAL r=32.}
\begin{tabular}{l|ccccccccc}
\toprule
\textbf{Rank} & \textbf{BoolQ} & \textbf{PIQA} & \textbf{SIQA} & \textbf{HellaSwag} & \textbf{Wino.} & \textbf{ARC-c} & \textbf{ARC-e} & \textbf{OBQA} & \textbf{Avg.} \\
\midrule
4 & 65.05 & 78.18 & 75.64 & 76.16 & 73.56 & 65.02 & 81.65 & 74.80 & 73.76 \\
8 & 64.83 & 81.23 & 77.02 & 83.92 & 77.35 & 68.43 & 83.00 & 79.20 & 76.87 \\
16 & 66.24 & 82.32 & 77.94 & 86.10 & 79.24 & 67.32 & 83.12 & 78.60 & \textbf{77.61} \\
32 & 66.45 & 82.16 & 78.20 & 83.72 & 79.95 & 68.09 & 82.62 & 79.40 & 77.57 \\
\midrule
$\text{LoRA}_{r = 32}$ & 65.96 & 78.62 & 75.23 & 79.20 & 76.64 & 79.13 & 62.80 & 72.40 & 73.75 \\
\bottomrule
\end{tabular}
\label{tab:rank_ablation_study}
\end{table*}

\subsection{Impact of the Size of Passport \(C\)}

\newcommand{\std}{\texttt{std}}
\label{sec:passport_std_appendix}

\begin{figure*}[!ht]   
    \caption{Effect of passport $C$ standard deviation (\texttt{std}) on \scheme{} weight. \texttt{std} = $\sigma$: Outputs are using only SEAL weight without \(C \sim \mathcal{N}(0, \sigma^2) \), $\mathbb{N}(B, A, \emptyset)$. Vanilla SD 1.5: output from vanila Stable Diffusion 1.5 with same prompt.}
    \centering
    \includegraphics[width=\columnwidth]{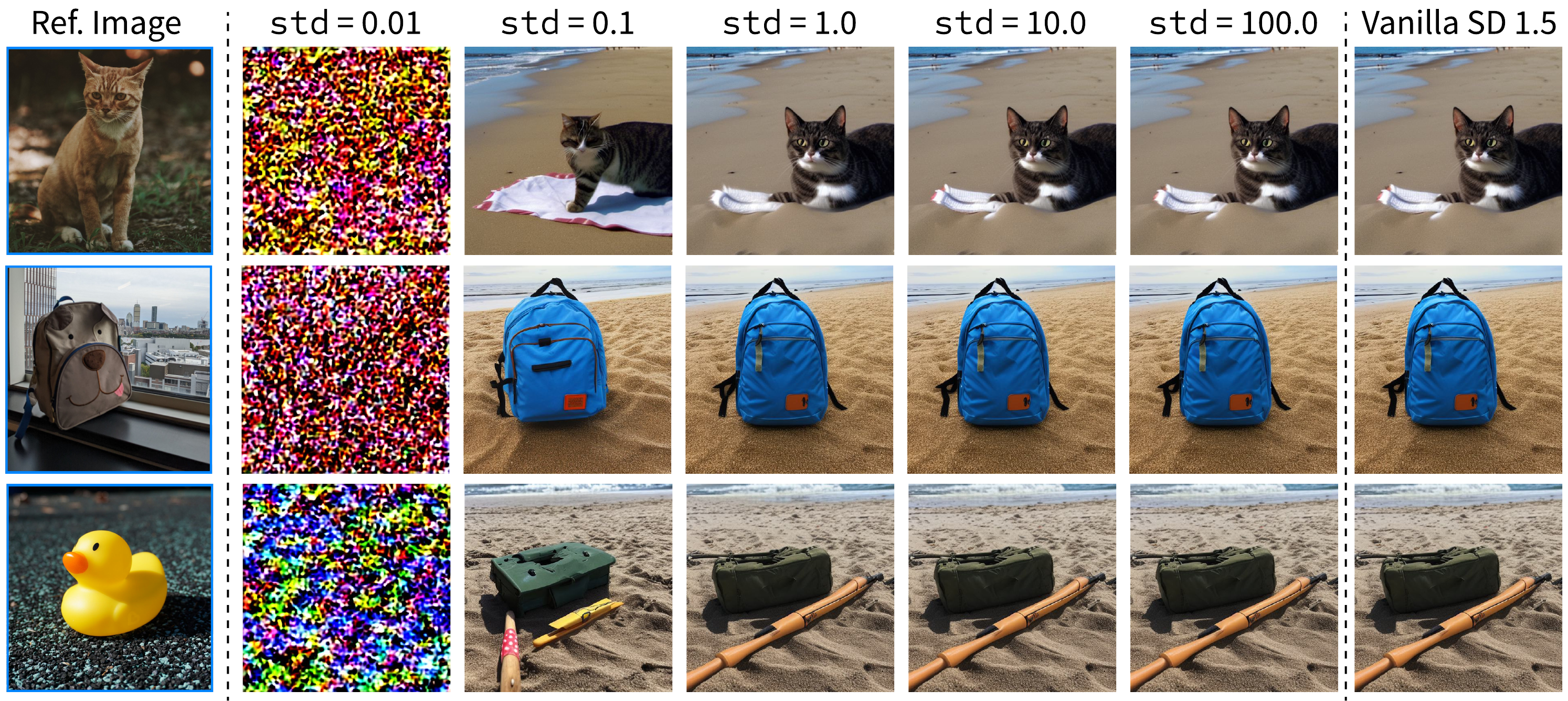}
    \label{fig:dreambooth_std_comparison}
\end{figure*}

To analyze how the magnitude of the passport \(C\) influences the final output, 
we train the model with \(\Delta W = B\,C\,A\), but at inference time remove \(C\) 
(i.e., \(\mathbb{N}(B, A, \emptyset)\)) to observe the resulting images 
under different standard deviations \(\std{}\) of \(C\). 
Specifically, we sample \(C \sim \mathcal{N}(0, \std^2)\) with \(\std \in \{0.01,0.1,1.0,10.0,100.0\}\) 
and keep \(B\) and \(A\) trainable. 
\figref{fig:dreambooth_std_comparison} shows that lower \(\std\) 
(e.g., \(0.01\)) produces markedly different images relative to the vanilla model 
\textbf{without} $C$, while higher \(\std\) (e.g., \(10.0\) or \(100.0\)) 
yields outputs closer to the vanilla Stable Diffusion model\footnote{\url{https://huggingface.co/stable-diffusion-v1-5/stable-diffusion-v1-5}. The original weight had been taken down.}.

\paragraph{Why does \(\std\) of \(C\) affect \(\mathbb{N}(B,A,\emptyset)\)?}
Recall that \(\Delta W = B\,C\,A\). 
If \(\std(C)\) is very small (e.g., \(0.01\)), then during training, 
the product \(B\,C\,A\) must still approximate the desired update \(\Delta W\). 
Because \(C\) is tiny, \(B\) and \(A\) tend to have relatively large values 
to compensate. 
Consequently, when we \emph{remove} \(C\) at inference time 
(use \(\mathbb{N}(B,A,\emptyset)\)), these enlarged \(B\) and \(A\) 
inject strong perturbations, manifesting visually as high-frequency artifacts.  

Conversely, if \(\std(C)\) is very large (e.g., \(10.0\) or \(100.0\)), 
then to avoid destabilizing training, \(B\) and \(A\) remain smaller in scale. 
Hence, removing \(C\) at inference, \(\mathbb{N}(B,A,\emptyset)\), 
introduces only minor differences from the original model, 
leading to outputs that closely resemble the vanilla Stable Diffusion model.  

\paragraph{Quantitative Comparison.}
In addition to the qualitative results, \tabref{tab:psnr_ssim_comparison} compares 
Peak Signal-to-Noise Ratio (PSNR) and Structural Similarity (SSIM) 
between images generated 
using only trained \scheme{} weights \textbf{without} $C$, $\mathbb{N}(B, A, \emptyset)$, at various passport \(\std\) values. 
Lower \(\std\) (e.g., \(0.01\)) shows significantly lower PSNR and SSIM, 
indicating large deviations (i.e., stronger perturbations) from the vanilla output. 
As \(\std\) increases to \(10.0\) or \(100.0\), the outputs become more aligned 
with the vanilla model, reflected by higher PSNR/SSIM scores.  

\begin{table}[!ht]
    \centering
    \caption{%
    Comparision of PSNR and SSIM values for images generated \textbf{without} \(C \sim \mathcal{N}(0, \sigma^2) \), using only $\mathbb{N}(B, A, \emptyset)$,
    under varying standard deviations of the passport \(C\), with images generated under vanilla SD 1.5 model.  
    Obj.~1: \texttt{Cat}, Object~2: \texttt{Backpack dog}, Obj.~3: \texttt{Ducky toy}. Object names are same as~\cite{ruiz2023dreambooth}
    }
    \setlength{\tabcolsep}{9pt} 
    \renewcommand{\arraystretch}{1.1}
    \begin{tabular}{l|l|ccccc}
        \toprule
        \multirow{2}{*}{\textbf{Ref.}} 
        & \multirow{2}{*}{\textbf{Metric} $\uparrow$} 
        & \multicolumn{5}{c}{\textbf{Standard Deviation of }$C$} \\
        & & \textbf{0.01} & \textbf{0.1} & \textbf{1.0} & \textbf{10.0} & \textbf{100.0} \\
        \midrule
        \multirow{2}{*}{Obj.1} 
        & SSIM & 0.104 & 0.691 & 0.936 & 0.987 & 0.998 \\
        & PSNR & 7.80  & 19.02 & 30.87 & 43.64 & 53.16 \\
        \midrule
        \multirow{2}{*}{Obj.2} 
        & SSIM & 0.102 & 0.652 & 0.941 & 0.993 & 0.998 \\
        & PSNR & 7.91  & 18.51 & 33.15 & 47.24 & 54.21 \\
        \midrule
        \multirow{2}{*}{Obj.3} 
        & SSIM & 0.115 & 0.651 & 0.959 & 0.992 & 0.998 \\
        & PSNR & 8.08  & 18.39 & 32.92 & 45.39 & 53.58 \\
        \bottomrule
    \end{tabular}
    \label{tab:psnr_ssim_comparison}
\end{table}

\section{Extending to Multiple Passports and Data-based Mappings}
\label{appendix:multi_passports}

So far, our main exposition has treated the watermark matrices $C$ and $C_p$, constant passports.
However, \scheme{} naturally extends to a setting in which one maintains multiple passports $\{C_1, C_2, \ldots, C_m\}$ (similarly $\{D_1, D_2, \ldots\, D_n$),
each possibly tied to a distinct portion of the training set, or to a distinct sub-task within the same model.
Formally, suppose that during mini-batch updates \algref{alg:training} randomly picks \emph{one} passport $C_i$ \emph{associated} with $(x,y)$.
Then line 10 of~\algref{alg:training} becomes:
\[
    \text{pick } C_i \;\text{s.t.\ } (x,y) \;\mapsto\; C_i, 
    \quad
    W' \;\leftarrow\; W + B\,C_i\,A.
\]
One can store a simple mapping function $\phi:\,(x,y)\!\mapsto i \in \{1,\ldots,m\}$ to tie each batch to its specific passport.

\textbf{Distributed or Output-based Scenarios.}
Another angle is to use multiple passports not only at \emph{training} time but also during \emph{inference}. 
For instance, given a family $\{C_1,\dots,C_m\}$, one could selectively load $C_i$ to induce different behaviors or tasks in an otherwise single LoRA model. 
In principle, if each $C_i$ is entangled with $(B,A)$, switching passports at inference changes the effective subspace.
This may be viewed as a \emph{distributed watermark} approach: where each $C_i$ can be interpreted as a unique “key” that enables (or modifies) certain model capabilities, separate from the main training objective.
Though we do not explore this direction in detail here, it points to broader usage possibilities beyond simply verifying ownership, such as controlled multi-task inferences and individually licensed feature sets.

\end{document}